\documentclass[sigconf,nonacm]{acmart}

\def\BibTeX{{\rm B\kern-.05em{\sc i\kern-.025em b}\kern-.08emT\kern-.1667em\lower.7ex\hbox{E}\kern-.125emX}}

\usepackage{setspace}
\usepackage{bm}
\usepackage{xcolor}
\usepackage{booktabs}
\usepackage{microtype}
\usepackage{amsfonts}
\usepackage{xspace}
\usepackage{wasysym}
\usepackage{enumitem,etoolbox}
\usepackage{array}
\usepackage{microtype}
\usepackage{multirow}
\usepackage{framed}
\usepackage{caption}
\usepackage{threeparttable}
\usepackage{appendix}
\usepackage[labelformat=simple]{subcaption}

\usepackage[ruled,vlined]{algorithm2e}
\usepackage[normalem]{ulem}

\usepackage{xurl}

\SetCommentSty{mycommfont}

\SetKwInOut{Input}{Input}
\SetKwInOut{Output}{Output}
\SetKwInOut{Except}{Except.}
\SetKw{None}{nil}
\SetKw{Cont}{continue}
\SetKw{Break}{break}
\SetKwProg{Fn}{Function}{:}{}
\SetKwFunction{Initp}{InitPattern}
\SetKwFunction{Nextp}{NextPattern}
\SetKwFunction{Initss}{InitSerSched}
\SetKwFunction{Nextss}{NextSerSched}
\SetAlCapFnt{\small}
\SetAlCapNameFnt{\small}
\LinesNumbered
\DontPrintSemicolon

\usepackage{tikz}
\usetikzlibrary{positioning}
\tikzstyle{period} = [draw=white, fill=gray!30, thick,
    rectangle, rounded corners, minimum size=9ex]
\usepackage{subcaption}
\usepackage{soul}

\AtBeginEnvironment{verbatim}{\setlist[trivlist]{nosep}}
\usepackage{flushend}
\usepackage{balance}
\usepackage{graphicx}
\usepackage{amsthm}
\usepackage{amsmath}
\usepackage{amssymb}

\usepackage{bbding}
\usepackage{array}
\usepackage{multirow}
\usepackage{listings}
\usepackage{color}
\usepackage{tcolorbox}
\usepackage{xcolor,colortbl}
\usepackage{hyperref}
\usepackage{enumitem}
\usepackage{adjustbox}
\usepackage[normalem]{ulem}
\usepackage{pifont}

\definecolor{bg}{HTML}{F8F9FB}  
\definecolor{rowcolor}{HTML}{ECEFF4}
\definecolor{greenbg}{HTML}{D2E4DB}
\definecolor{added}{HTML}{C7E0D6}
\definecolor{removed}{HTML}{FBDBD8}

\usetikzlibrary{shapes,positioning}
\makeatletter
\let\old@lstKV@SwitchCases\lstKV@SwitchCases
\def\lstKV@SwitchCases#1#2#3{}
\makeatother
\usepackage{lstlinebgrd}
\makeatletter
\let\lstKV@SwitchCases\old@lstKV@SwitchCases

\lst@Key{numbers}{none}{%
    \def\lst@PlaceNumber{\lst@linebgrd}%
    \lstKV@SwitchCases{#1}%
    {none:\\%
        left:\def\lst@PlaceNumber{\llap{\normalfont
                \lst@numberstyle{\thelstnumber}\kern\lst@numbersep}\lst@linebgrd}\\%
        right:\def\lst@PlaceNumber{\rlap{\normalfont
                \kern\linewidth \kern\lst@numbersep
                \lst@numberstyle{\thelstnumber}}\lst@linebgrd}%
    }{\PackageError{Listings}{Numbers #1 unknown}\@ehc}}
\makeatother

\setlist[itemize]{leftmargin=*}

\definecolor{dkgreen}{rgb}{0,0.6,0}
\definecolor{gray}{rgb}{0.5,0.5,0.5}
\definecolor{mauve}{rgb}{0.58,0,0.82}

\newcommand{\eg}{\hbox{\emph{e.g.}}\xspace}
\newcommand{\ie}{\hbox{\emph{i.e.}}\xspace}
\newcommand{\etc}{\hbox{\emph{etc.}}\xspace}

\newcommand{\completek}{Completion@$k$\xspace}
\newcommand{\compilek}{Compilation@$k$\xspace}
\newcommand{\testpassk}{Pass@$k$\xspace}

\setlist[itemize]{leftmargin=*}
\setlist[enumerate]{leftmargin=*}
\newlist{steps}{enumerate}{1}
\setlist[steps, 1]{label = \textbf{RQ\arabic*.}}

\definecolor{Gray}{gray}{0.9}

\newcommand{\name}{{JavaBench}\xspace}

\definecolor{bg}{HTML}{F8F9FB}  
\usepackage[framemethod=TikZ]{mdframed}
\mdfdefinestyle{MyFrame}{%
    linecolor=black,
    outerlinewidth=0.3pt,
    roundcorner=5pt,
    skipabove = 5.5pt,
    skipbelow = 5.5pt,
    innertopmargin=5.5pt, 
    innerbottommargin=5.5pt, 
    innerrightmargin=5.5pt,
    innerleftmargin=5.5pt,
    backgroundcolor=bg, 
}

\begin{document}

\title[\name]{\name: A Benchmark of Object-Oriented Code Generation for Evaluating Large Language Models}

\author{Jialun Cao}
\authornotemark[1]
\affiliation{
  \institution{Department of Computer Science \\and Engineering, The Hong Kong University of Science and Technology, Hong Kong, China
  \\Guangzhou HKUST Fok Ying Tung Research Institute, Guangzhou, China}
  }

\author{Zhiyong Chen}
\authornotemark[1]
\thanks{{*} Co-first authors. }
\affiliation{
  \institution{State Key Laboratory for Novel Software Technology, Nanjing University}
  \city{Nanjing}
  \country{China}}

\author{Jiarong Wu}
\affiliation{
  \institution{Department of Computer Science \\and Engineering, The Hong Kong University of Science and Technology, Hong Kong, China
  \\Guangzhou HKUST Fok Ying Tung Research Institute, Guangzhou, China}
  }

\author{Shing-Chi Cheung}
\affiliation{
  \institution{Department of Computer Science \\and Engineering, The Hong Kong University of Science and Technology, Hong Kong, China
  \\Guangzhou HKUST Fok Ying Tung Research Institute, Guangzhou, China}
  }

\author{Chang Xu}
\affiliation{
  \institution{State Key Laboratory for Novel Software Technology, Nanjing University}
  \city{Nanjing}
  \country{China}}

%
\renewcommand{\shortauthors}{Jialun Cao and Zhiyong Chen et al.}

\begin{abstract}
Code generation benchmarks such as HumanEval are widely adopted to evaluate LLMs' capabilities.
However, after consolidating the latest 24 benchmarks, we noticed three significant imbalances.
First, \textbf{\textit{imbalanced programming language}}.
95.8\% of benchmarks involve Python, while only 5 benchmarks involve Java, resulting in an insufficient understanding of LLMs' capability to generate Java code.
Second, \textbf{\textit{imbalanced code granularity}}.
Function-/statement-level benchmarks account for over 83.3\% of benchmarks.
Only a mere handful extends to class-/project-levels, and all are limited to Python. 
Third, \textbf{\textit{lacking advanced features}}.
Existing benchmarks primarily assess basic coding skills (\eg, variables, operators, and control structures), while overlooking advanced Object-Oriented Programming (OOP) features (\ie, encapsulation, inheritance, and polymorphism).  
Considering the prevalence of these advanced features in real-world Java project development, constructing benchmarks to test LLMs on handling OOP features is necessary.

To fill these gaps, we propose \textbf{\textit{\name, a project-level Java benchmark that exercises OOP features}}.
It comprises four Java projects with 389 methods in 106 Java classes.
The test coverage is up to 92\%, and \name is attested by 282 undergraduate students, reaching a 90.93/100 average score (\ie, pass rate against the test suite), ensuring the quality of documentation, code skeleton, and tests.
To better evaluate LLM's capability against \name, we introduce a systematic evaluation design covering three context settings and five synthesis strategies at two granularities using three hierarchical metrics. 
Our extensive experiment yields several interesting findings.
First, we noticed that regarding project-level Java programming, \textbf{\textit{LLMs are far behind undergraduate students}} (\textbf{\textit{no project}} can be correctly completed by any studied LLMs, and at most \textbf{\textit{41.17\%}} Pass@5 in a more relaxed evaluation).
Second, {{using method signature as prompt context may strike an ideal balance}} for project-level code generation. 
{{\name}} is publicly \textbf{\textit{available}} at \url{https://github.com/java-bench/JavaBench}.
We also release a \textbf{\textit{leaderboard}} and invite model developers to participate and test their models against \name at \url{https://java-bench.github.io/leaderboard.html}.

\end{abstract}

%
%

%


%
\settopmatter{printfolios=true}
\maketitle



\section{Introduction}\label{sec:intro}
Large language models (LLMs) such as ChatGPT~\cite{gpt35turbo,gpt4turbo} have shown advanced proficiency~\cite{belzner2023large} in various tasks such as code generation~\cite{humaneval,du2023classeval,yu2023codereval}, code reasoning~\cite{gu2024cruxeval} and code summarization~\cite{gao2023codesummary}.
Emerging \textbf{\textit{code generation/completion benchmarks}}~\cite{humaneval,yu2023codereval,mbpp2021,concode,humanevalX,ding2023crosscodeeval,devEval,repoeval} like HumanEval~\cite{humaneval} have been introduced to evaluate LLMs' capabilities, providing insights into their strengths and weaknesses in various {\textit{real-world}} scenarios, thereby guiding LLM researchers to address related issues more effectively.

\textbf{\textit{Research gap}} -- 
To gain a comprehensive overview of the current state of these benchmarks, we consolidated the data from the recent studies~\cite{wang2024oop,nl2code-survey,zheng2023survey} and incorporated the latest benchmarks, resulting in Table~\ref{tab:benchmarks}.
By analyzing the statistics, we identified three significant imbalances.
\textbf{\textit{1. Imbalanced Programming Languages}. }
There is a disproportionate focus on Python, which constitutes 95.8\% (23/24) of benchmarks.
Java, despite being the second most popular language on GitHub~\cite{github-stats} (Java holds 11.708\% while Python holds 16.925\% and is ranked first), is covered by only five function-level benchmarks.
The lack of Java benchmarks limits the understanding of LLMs' capabilities in generating Java code compared to Python.
\textbf{\textit{2. Imbalanced Code Granularity}. }
These benchmarks predominantly feature function-level granularity or lower (\ie, statement-level), accounting for 83.3\% (20/24) of the total. 
Although these benchmarks can exercise LLMs' ability to generate code for individual functions, a broader context (\eg, cross-function/class) is often required in real-world development scenarios, \eg, inheriting a class and overwriting the interface~\cite{devEval}.
Such scenarios cannot be adequately assessed by statement-/function-level benchmarks. 
Only a mere handful extends to class- or project-levels, and all are limited to Python. 
\textbf{\textit{3. Lacking Advanced Features}.}
Current benchmarks comprehensively assess basic coding skills (\eg, variables, data types, operators, and control structures) while overlooking advanced {\textit{Object-Oriented Programming (OOP) features}} (encapsulation, inheritance, and polymorphism). 
OOP promotes modularity and reusability of the code and is thus commonly adopted in real-world development. 
However, only one recent benchmark~\cite{wang2024oop} claims to exercise OOP features, and it does not provide actual code context but merely mentions the OOP concept in the prompt.
In summary, {{there is a clear gap to fill to adequately test LLMs in handling OOP features}}, motivating the need for more comprehensive benchmarks.

\textbf{Benchmark \name} -- 
To bridge the research gap, we propose \name, a \textbf{\textit{project-level Java benchmark that exercises OOP features}} (\ie, encapsulation, inheritance, and polymorphism). 
It comprises four Java projects that were programming assignments in an entry-level Java course.
These four projects contain 389 methods in 106 Java classes, covered by 396 tests, reaching up to 92\% code coverage.
In addition, \name is attested by 282 undergraduate students, reaching a 90.93/100 average score (\ie, pass rate against the test suite), ensuring the quality of documentation, code skeleton, and tests in \name.
Furthermore, we extensively evaluate \textbf{\textit{five LLMs}} (\eg, gpt-3.5, DeepSeeker, Phind) against \name under a set of comprehensive settings.
In particular, we design \textbf{\textit{three context settings}}  (\ie, maximum/minimum/selected context) in prompting, adopt \textbf{\textit{five synthesis strategies}} (\ie, holistic, independent, incremental, and its two variants), and evaluate the synthesized projects at \textbf{\textit{two evaluation granularities}} (\ie, class-wise and test-wise) using \textbf{\textit{three metrics}} (\ie, \completek, \compilek, and \testpassk).

Our extensive experiments yield several interesting findings. 
First, in terms of project-level Java programming ability, \textbf{\textit{LLMs are still far behind undergraduate students}}. 
The best LLMs under the best setting only reach a 41.7\% Pass@5 in test-wise granularity (Section~\ref{sec:rq1}), compared with 90.93\% achieved by undergraduate students under a stricter evaluation. 
Second, Providing \textbf{\textit{method signature only}} in the prompt leads to optimal results, while too much or too little context degrades project-level code generation.  

\vspace{2pt}
\noindent\textbf{\textit{Contributions --}} 
Our contribution is summarized as follows. 

\begin{itemize}
    \item \textit{\textbf{Significance.} }
    We proposed the first project-level Java benchmark that exercises OOP features (i.e., encapsulation, inheritance, and polymorphism). It enables observations of LLMs' strengths and weaknesses in handling Java OOP features.
    
    \item \textit{\textbf{Novelty.} }
    Besides introducing \name, we also introduce a systematic evaluation design to assess LLMs' capabilities under three context settings at two evaluation granularities using three progressive metrics. This evaluation design provides a reference for future project-level code generation assessments.
        
    \item \textit{\textbf{Evaluation.}} 
    We conduct extensive experiments that yield several instructive findings. We point out that LLM's capability to handle OOP features is far behind that of undergraduates. 
    We also identified an optimal context setting with only method signatures provided. Our analysis of bad cases also provides directions for future improvement.
    
\end{itemize}

\section{Benchmark Construction}\label{sec:benchmark}

\subsection{Benchmark Format}\label{sec:format}
\begin{figure}[t!]
    \centering
    \includegraphics[width=1.0\linewidth]{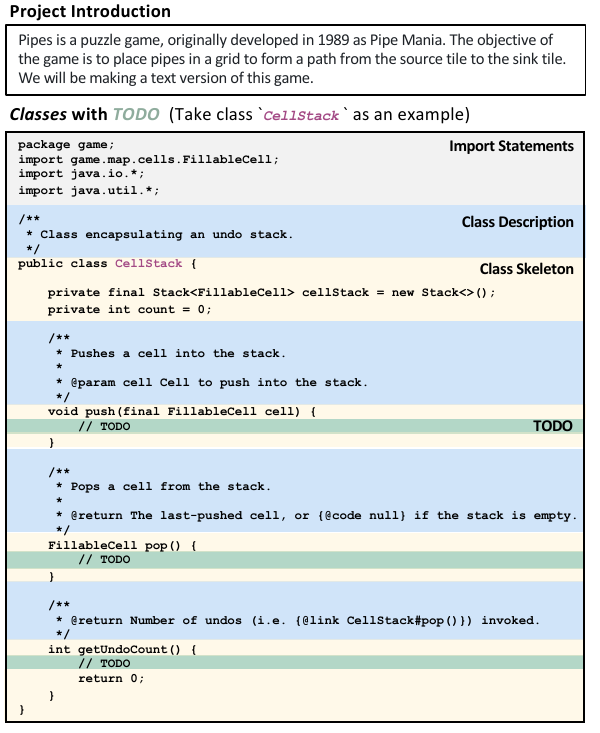}
    \setlength{\abovecaptionskip}{-0pt}
    \setlength{\belowcaptionskip}{-12pt}
    \caption{An Example of Project Skeleton in \name}
    \label{fig:skeleton}
\end{figure}

An example of a Java project in \name is illustrated in Figure~\ref{fig:skeleton}.
A project comprises a description of the whole project in natural language and a \textbf{\textit{code skeleton}} with multiple classes (Figure~\ref{fig:skeleton} only shows one class due to space limit).
Each class includes import statements, a class description, a class skeleton with multiple methods.
Each method has a docstring and can be complete or incomplete, \ie, the method body is a \texttt{TODO} to be filled in by LLMs.

\begin{table}[t!]
\centering
\caption{Summarization of 24 Existing Benchmarks plus \name.
The ones involving Java are highlighted in \colorbox{Gray}{gray}.
}
\label{tab:benchmarks}
\renewcommand\arraystretch{1.2}
    \resizebox{1.0\linewidth}{!}{
\begin{tabular}{l|lllrrrrr}
\toprule
Benchmark & Time & Language & Granularity & \# Funcs & \# Class & \#  AvgT & \# Tests & \# AvgLOC \\
\midrule
\rowcolor{Gray}
Concode~\cite{concode} & 2018 & \textbf{Java} & Function & 2,000 & 0 & - & - &  \\
CoNaLA\cite{yin2018mining} & 2018 & Python & Statement & 500 & 0 & - & - & 1 \\
APPS\cite{hendrycksapps2021} & 2021 & Python & Function & 5,000 & 0 & 13.2 & 66,000 & 21.4  \\
HumanEval~\cite{humaneval} & 2021 & Python & Function & 164 & 0 & 7.7 & 1,263 & 11.5 \\
MBPP~\cite{mbpp2021} & 2021 & Python & Function & 974 & 0 & 3 & 2,922 & 6.8  \\
math-qa~\cite{mathqa2019} & 2021 & Python & Statement & 2,985 & 0 & - & - & 7.6 \\
\rowcolor{Gray}
HumanEval-X~\cite{humanevalX} & 2022 & Python, \textbf{Java}, \etc & Function & 164 & 0 & 7.8 & 1,279 & 12.1 \\
\rowcolor{Gray}
MBXP~\cite{mbxp_athiwaratkun2022} & 2022 & Python, \textbf{Java}, \etc & Function & 974 & 0 & 3 & 2,922 & 6.8  \\
CodeContests & 2022 & Python, C++ & Function & 165 & 0 & 203.7 & 33,610 & 59.8 \\
PandasEval\cite{zan2022cert} & 2022 & Python & Function & 101 & 0 & 6.5 & 656 & 1.3 \\
NumpyEval\cite{zan2022cert} & 2022 & Python & Function & 101 & 0 & 3.5 & 354 & 1.1 \\
TorchDataEval\cite{zan2022language} & 2022 & Python & Function & 50 & 0 & 1.1 & 55 & 1.3 \\ 
DS-1000~\cite{Lai2023DS1000} & 2022 & Python & Statement & 1,000 & 0 & 1.6 & 1,600 & 3.8 \\
DSP\cite{chandel2022training} & 2022 & Python & Function & 1,119 & 0 & 2.1 & 2,350 & 7.6 \\
\rowcolor{Gray}
MultiPL-MBPP\cite{cassano2022scalable} & 2022 & Python, \textbf{Java}, \etc & Function & 974 & 0 & 3.1 & 3,019 & - \\
MTBP\cite{nijkamp2022codegen} & 2022 & Python & Function & 115 & 0 & - & - & - \\
ODEX\cite{wang2022execution} & 2022 & Python  & Function & 945 & 0 & 1.8 & 1,701 & 1.9 \\
BIG-Bench\cite{srivastava2023beyond} & 2023 & Python & Function & 32 & 0 & 4.7 & 150 & - \\
\rowcolor{Gray}
CoderEval~\cite{yu2023codereval} & 2023 & Python, \textbf{Java} & Function & 230+230 & 0 & - & - & $\leq 32$ \\
\rowcolor{Gray}
CrossCodeEval~\cite{ding2023crosscodeeval} & 2023 & Python, \textbf{Java}, \etc & Statement & - & 3,534 & 0 & 0 & 96.2 \\
RepoEval~\cite{repoeval} & 2023 & Python & \textbf{Project} & 1,973 & 0 & - & - & $\leq 30$ \\
\rowcolor{Gray}
ClassEval~\cite{du2023classeval} & 2023 & Python, \textbf{Java}, \etc & Class & 412 & 100 & 33.1 & 3,310 & 45.7  \\
DevEval~\cite{devEval} & 2024 & Python & \textbf{Project} & 1,874 & 0 & - & - & 392.7  \\
OOPEval\cite{wang2024oop} & 2024 & Python & \textbf{Project} & 0 & 431 & 2.5 & 1,070 & 0  \\
\midrule
\textbf{\name} & 2024 & \textbf{Java} & \textbf{Project} & 389 & 106 & 99 & 396 & 1,740\\
\bottomrule
\end{tabular}
}
\end{table}

\begin{table*}[th]
\centering
\caption{Summary of \name Projects}
\label{tab:pa-intro}
\renewcommand\arraystretch{1.0}
    \resizebox{1.0\textwidth}{!}{
\begin{tabular}{l|l|l|rrr}
\toprule 
\multirow{3}{*}{\textbf{ID}} & \multicolumn{1}{c|}{\multirow{3}{*}{\textbf{Description}}} & \multicolumn{1}{c|}{\multirow{3}{*}{\textbf{Exercised Concepts}}} & \multicolumn{3}{c}{\textbf{Human Performance}} \\ \cline{4-6} 
 & \multicolumn{1}{c|}{} & \multicolumn{1}{c|}{} & \multicolumn{1}{c|}{\multirow{2}{*}{\textbf{\# Stu}}} & \multicolumn{1}{c|}{\multirow{2}{*}{\textbf{Min$\sim$Max}}} & \multicolumn{1}{c}{\multirow{2}{*}{\textbf{Mean $\sim$$\pm$$\sim$Std}}} \\
 & \multicolumn{1}{c|}{} & \multicolumn{1}{c|}{} & \multicolumn{1}{c|}{} & \multicolumn{1}{c|}{} & \multicolumn{1}{c}{} \\ \hline
P1 & \begin{tabular}[c]{@{}l@{}}The project is a text-based version of \textbf{Pipe Mania} using Java. The game involves placing \\ pipes on a grid to connect a source to a sink, utilizing ASCII and Unicode for visualization. \\ Features include interactive controls, water flow simulation, and strategic game-play with \\ conditions for winning and losing.\end{tabular} 

& \begin{tabular}[l]{@{}l@{}}Basic Java, Interface, \textbf{Encapsulation}, \textbf{Inheritance}, \\  Overriding, \textbf{Polymorphism}, \underline{File IO}, Exception Handling\end{tabular} 
& \multicolumn{1}{|r|}{62} & \multicolumn{1}{r|}{52.88$\sim$100} & $\textbf{95.41}~\pm~7.26$ \\ \hline

P2 & \begin{tabular}[c]{@{}l@{}}The project is a text-based console version of \textbf{Jeson Mor}, a Mongolian strategy board game. \\Using Java, students will implement game mechanics where two players use knights, similar \\ to those in chess, to compete by capturing a central square on the board.\end{tabular} 

& \begin{tabular}[c|]{@{}l@{}}Basic Java, \underline{Streams}, \textbf{Encapsulation}, \textbf{Inheritance}, \\ \textbf{Overriding}, \textbf{Polymorphism} \\ Exception Handling\end{tabular} 
& \multicolumn{1}{|r|}{64} & \multicolumn{1}{r|}{20.67$\sim$100} & $\textbf{91.73}~\pm~15.05$ \\ \hline

P3 & \begin{tabular}[c]{@{}l@{}}The project is an ASCII version of the \textbf{Inertia puzzle game} in Java. The game challenges \\ players to navigate a board to collect gems while avoiding mines, with movement continuing \\ in one direction until an obstruction is encountered.\end{tabular} & \begin{tabular}[c]{@{}l@{}}Basic Java, \textbf{Encapsulation}, \textbf{Inheritance}, \\ \textbf{Overriding}, \textbf{Polymorphism}, \\ Exception Handling\end{tabular} & \multicolumn{1}{|r|}{77} & \multicolumn{1}{r|}{26.79 $\sim$100} & $\textbf{90.39}~\pm~16.66$ \\ \hline

P4 & \begin{tabular}[c]{@{}l@{}}The project is a modified \textbf{Sokoban game} featuring a text-based user interface. This enhanced \\ version introduces multiplayer functionality, allowing several players to simultaneously \\ navigate and manipulate designated boxes toward specific locations on the game map. \\ \end{tabular} & \begin{tabular}[c]{@{}l@{}}Basic Java, \textbf{Encapsulation}, \textbf{Inheritance}, \\ \textbf{Overriding}, \textbf{Polymorphism}, \underline{Mocking}, \\ Exception Handling, \underline{Streams}, \underline{Regex}\end{tabular} & \multicolumn{1}{|r|}{79} & \multicolumn{1}{r|}{34.27 $\sim$100} & $\textbf{90.96}~\pm~14.03$ \\ \hline
 & \multicolumn{1}{r|}{\textbf{Total}} &  & \multicolumn{1}{|r|}{\textbf{282}} & \multicolumn{1}{r|}{20.67 $\sim$ 100} & \textbf{90.93}$~\pm~$14.05 
 \\ 
\bottomrule
\end{tabular}
}
\end{table*}

\subsection{Benchmark Specification}\label{sec:construct}
We describe \name from the following three perspectives: 
(1) {Project Description} (Section~\ref{sec:proj}) describes the projects in \name and the corresponding Java features they exercised. 
(2) {Test Construction} (Section~\ref{sec:test}) describes the process of constructing test cases and reports the code coverage. 
(3) {Human Performance} (Section~\ref{sec:human}) shows how first- and second-year undergraduates perform in these projects of \name. 

\subsubsection{\textbf{Project Description}}\label{sec:proj}
A summary of the four Java projects in \name is given in Table~\ref{tab:pa-intro}.
The primary design goal behind designing these student projects is to craft exciting and engaging Java projects (\eg, chess games) encompassing a broad array of Java features, including basic Java functionalities, advanced object-oriented programming concepts (\eg, inheritance, polymorphism), and other skills such as file reading and exception handling for undergraduates to practice Java programming.
Each project covers similar Java concepts, with a slight variant highlighted by \underline{underscore}.
Such a design goal also fits the benchmarking of LLMs' capability to understand and exercise various Java features.
In particular, each project in \name is designed to exercise \textbf{\textit{OOP-related features}} (\ie, {inheritance}, {encapsulation}, and {polymorphism}), highlighted in bold in Table~\ref{tab:pa-intro}.

Besides, each project in \name has a canonical solution prepared by an experienced Java programmer with more than 5 years of experience and cross-validated by other experienced programmers. 
Moreover, these canonical solutions are released to more than 200 undergraduates (see Section~\ref{sec:human}) for review, ensuring the solutions' correctness.
Students are required to keep the course assignments and canonical solutions confidential for academic integrity,
which reduces the data contamination~\cite{dataContamination2023} threat to our benchmark.

The \textbf{\textit{number of functions and classes}} in each project is listed in Table~\ref{tab:test-stats}.
The four projects have similar scales, with 89 to 125 functions spreading across 23 to 30 Java classes.
In total, there are 389 functions and 106 Java classes in \name. 
The lines of code (\ie, LoC) of the entire project range from 2,560 to 6,926, with an average of 3,873 lines.
Excluding the lines of test suites, the remaining lines of codes are 1,352 to 2,373, with an average of 1,740. 
Compared with the existing Java benchmarks at the function level (Table~\ref{tab:benchmarks}), \name involves a much larger context size (1,740 vs. 392.7) and poses new challenges to Java code generation.

Furthermore, to get a better understanding of \name, we measure the \textbf{\textit{code complexity}} using two metrics (\ie, cyclomatic~\cite{codeComplexity,codeMetrics} and cognitive~\cite{bieri1955cognitive} complexity) as evaluated in existing works~\cite{yu2023codereval,cao2024concerned}.
We omit the formulas due to space limits.
Conceptually, these two metrics consider the number of decision points or branches, the nesting levels, or the number of logical operators.
As shown in Table~\ref{tab:test-stats} at the ``\textit{Complexity}'' entry, the four projects share similar code complexity values, 
with P3 being relatively easier than others and P1 being relatively more complex.

\subsubsection{\textbf{Test Construction}}\label{sec:test}
The test suites for each project in \name are manually constructed.
Similar to canonical solution construction, the test suites are constructed by experienced Java programmers, ensuring the exercised concepts in each project are covered by at least one test case.
Specifically, the statistics of tests for each project are tabulated in Table~\ref{tab:test-stats}.
There are 396 tests in total and 49 to 222 tests in each project, with an average of 99 tests.
The total lines of code in the test suites range are 8,532, with 2,133 on average.
The test sufficiency is shown by three test coverage metrics (\ie, class coverage, function coverage, and line coverage). As shown in the last column of Table~\ref{tab:test-stats}, 92\% classes, 87\% functions, and 86.75\% lines are covered by the test suites on average. 

\begin{table}[t!]
\centering
\caption{Code and Test Statistics of \name}
\label{tab:test-stats}
\renewcommand\arraystretch{1.0}
\resizebox{1.0\linewidth}{!}{
\begin{tabular}{l|rr|rr|rr|rr|rrr}
\toprule
\multicolumn{1}{c|}{} & \multicolumn{1}{c}{} & \multicolumn{1}{c|}{} & \multicolumn{2}{c|}{\textbf{LoC}} & \multicolumn{2}{c|}{\textbf{Complexity}} & \multicolumn{2}{c|}{\textbf{Test Info}} & \multicolumn{3}{c}{\textbf{Test Coverage (\%)}} \\ \cline{4-12} 
\multicolumn{1}{c|}{\multirow{-2}{*}{\textbf{ID}}} & \multicolumn{1}{c}{\multirow{-2}{*}{\textbf{Func}}} & \multicolumn{1}{c|}{\multirow{-2}{*}{\textbf{Class}}} & \multicolumn{1}{c}{Total} & \multicolumn{1}{c|}{w/o Ts} & \multicolumn{1}{c|}{Cyc} & \multicolumn{1}{c|}{Cog} & \multicolumn{1}{c}{\# Tests} & \multicolumn{1}{c|}{LoC} & \multicolumn{1}{c}{Class} & \multicolumn{1}{c}{Func} & \multicolumn{1}{c}{Line} \\ \hline
P1 & 89 & 24 & 2,560 & 1,709 & \multicolumn{1}{r|}{18.70} & 19.90 & 55 & 851 & 91 & 89 & 81 \\
P2 & 102 & 23 & 3,223 & 1,524 & \multicolumn{1}{r|}{8.93} & 9.71 & 49 & 1,699 & 95 & 81 & 87 \\
P3 & 125 & 29 & 6,926 & 2,373 & \multicolumn{1}{r|}{12.50} & 9.21 & 222 & 4,553 & 100 & 85 & 87 \\
P4 & 73 & 30 & 2,781 & 1,352 & \multicolumn{1}{r|}{16.57} & 10.86 & 70 & 1,429 & 80 & 93 & 92 \\
\midrule
Total & 389 & 106 & 15,490 & 6,958 & \multicolumn{1}{r|}{14.18} & 12.42 & 396 & 8,532 & 92 & 87 & 86.75\\
\bottomrule
\end{tabular}
}
\end{table}

Notably, we embrace \textit{\textbf{mocking}}~\cite{alshahwan2010automock,galler2010automaticallymock,arcuri2014automatedmock,arcuri2017privatemock,MockSniffer20} into the test suite. It does not affect the code generation, while helping isolate the component under test from its dependencies and increases test stability. 
An example of using mocking to test the termination status of a Sokoban Game (\ie, P4) is shown in Listing~\ref{listing:mocking}.
Typically, we implement it with \texttt{mockito} (line 2) and mock three objects (\ie, \texttt{GameState}, \texttt{TerminalInputEngine} and \texttt{TerminalRenderingEngine}) in lines 7-9.
Then, the assertion checks whether the exception has been thrown in line 12.
Indeed, embracing mocking into test suite design is worth the endeavor,
as emphasized by a recent study~\cite{study-llm4testgen2024}.

\definecolor{stubbg}{HTML}{FCF3D5}
\begin{figure}[t!]
\begin{lstlisting}[
		language=java,
        % basicstyle=\linespread{0.8}\tiny,
        % baselinestretch=1.0,
		morekeywords={when, assertThrowsExactly, var, mock},
            aboveskip=-2pt,
            belowskip=-15pt,
		caption={\textbf{An example of mocking test in \name}},
		label={listing:mocking},
		escapechar=|,
		linebackgroundcolor = {\ifnum \value{lstnumber} > 6 \ifnum \value{lstnumber} < 10 \color{stubbg} \fi \fi},
		numbers=left
	]
import org.junit.jupiter.api.Assertions.assertThrowsExactly;
import org.mockito.Mockito.*;

class TerminalSokobanGameTest {
    @Test
    void testMoreThanTwoPlayers() {
        final var gameState = mock(GameState.class);
        final var inputEngine = mock(TerminalInputEngine.class);
        final var renderingEngine = mock(TerminalRenderingEngine.class);
        when(gameState.getAllPlayerPositions()).thenReturn(new HashSet<>(Arrays.asList(Position.of(1, 1), Position.of(1, 2), Position.of(1, 3))));

        assertThrowsExactly(IllegalArgumentException.class, () -> new TerminalSokobanGame(gameState, inputEngine, renderingEngine));
    }
}
\end{lstlisting}
\end{figure}

\begin{figure*}[t]
    \centering
    \includegraphics[width=1.0\linewidth]{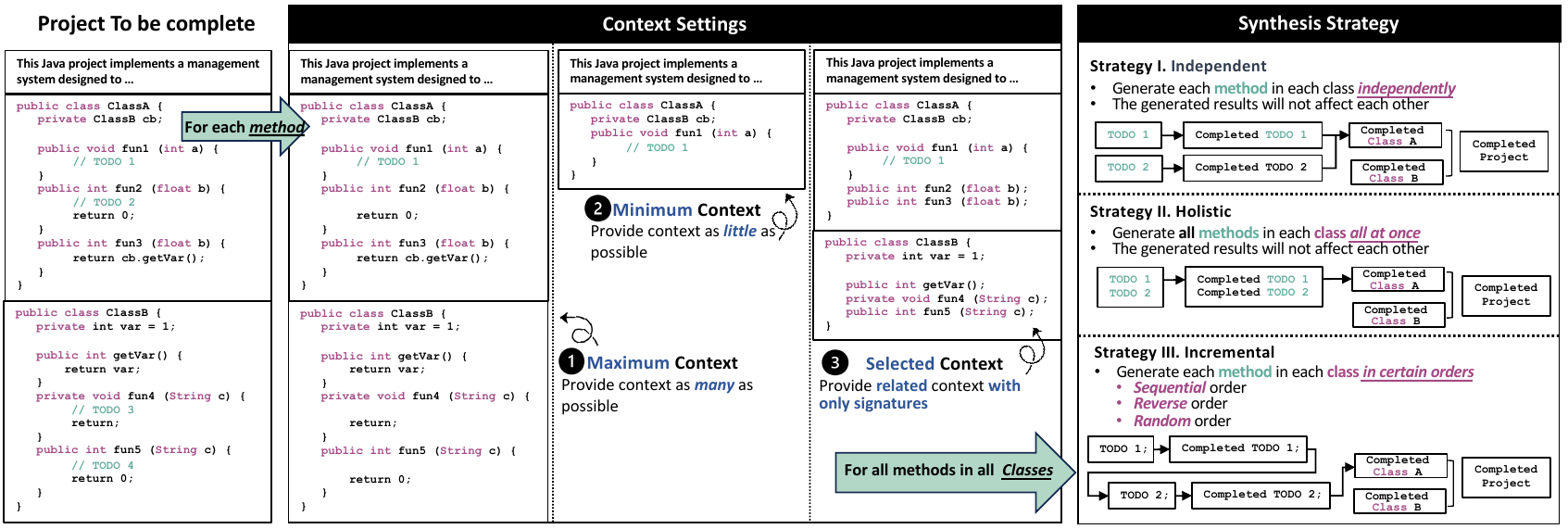}
    \setlength{\abovecaptionskip}{-0pt}
    \setlength{\belowcaptionskip}{-5pt}
    \caption{\textbf{Generation Pipeline for a Java Project.}  Given a project to be complete, for each \textit{method} with {\textcolor{teal}{\texttt{TODO}}}, there are three types of (\ding{202} $\sim$ \ding{204}) \textbf{Context Settings}. On top of method completion, there are three \textbf{Synthesis Strategies} to complete an entire \textit{class}. 
    }
    \label{fig:generate}
\end{figure*}

\subsubsection{\textbf{Human Performance}}\label{sec:human}

The four Java projects in \name are designed for undergraduate students throughout the four academic years from 2019 to 2022. 
We then use students' overall scores as indicators of difficulty levels.
As shown in the last entry ({Human Performance}) of Table~\ref{tab:pa-intro}, 282 undergraduates are involved~\footnote{When counting the number of participants, we omit course withdrawals, non-submissions, and blank project submissions from the count because these cases do not attempt to complete the project.},
and each project is finished by at least 62 students. 

To rate the students' submissions, we mainly use the \textbf{pass rates} (full score is 100) of the test suite as evidence.
The last two columns of Table~\ref{tab:pa-intro} demonstrate each project's maximum/minimum scores and mean and standard deviation.
The difficulty of all projects is similar, with an average of 90.96 to 95.41. 

\begin{mdframed}[style=MyFrame]
Undergraduates can finish the projects in \name with a 90.96\% to 95.41\% test pass rate (average \textbf{90.93\%}), indicating the difficulty of \name is acceptable for humans.
\end{mdframed}

\section{Experiment Design}\label{sec:design}

\begin{figure*}[t]
    \centering
    \includegraphics[width=1.0\linewidth]{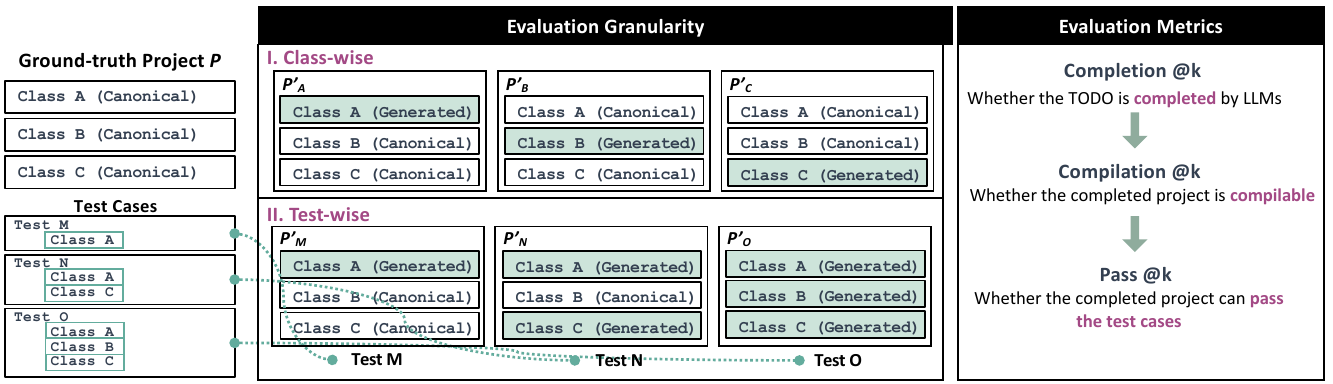}
    \setlength{\abovecaptionskip}{-0pt}
    \setlength{\belowcaptionskip}{-5pt}
    \caption{\textbf{Evaluation Design of Granularities and Metrics.} To evaluate an LLM-generated project, two \textbf{granularities} (\ie, class-wise and test-wise) are adopted to replace the related classes to compile corresponding programs $P'_X$ where $X$ denotes a class (A-C) or a test (M-O).
    Then, three-fold \textbf{evaluation metrics} (\ie, completion, compilation, and pass) are applied to evaluate $P'_X$.}
    \label{fig:evaluate}
\end{figure*}

Given a Java project skeleton, we first {{synthesize the entire project}} using three \textit{context settings} (Section~\ref{sec:context}) and various \textit{synthesis strategies} (Section~\ref{sec:syn}), as shown in Figure~\ref{fig:generate}. 
Then, we \textbf{\textit{evaluate}} the {{synthesized project}} in terms of three \textit{evaluation metrics} (Section~\ref{sec:metrics}) in two \textit{granularities} (Section~\ref{sec:granularity}), as shown in Figure~\ref{fig:evaluate}.

\subsection{\textbf{Code Synthesis}}
The synthesis pipeline is shown in Figure~\ref{fig:generate}. 
Given a skeleton project, we complete it class-by-class because a class is usually designed to be cohesive and less coupling with others.
To complete a method in a class, providing the context of a standalone method/class is insufficient due to the lack of dependencies between classes. 
Thus, we try three context settings (Section~\ref{sec:context}).
Since the order to generate multiple TODO methods in a class also matters, we try three synthesis strategies (Section~\ref{sec:syn}).
When all the methods in all the classes are synthesized, the skeleton project is completed.

\subsubsection{\textbf{Context Settings}}\label{sec:context}
We design three context settings in the synthesis pipeline, as shown in the middle part of Figure~\ref{fig:generate}. 
A straightforward context is to feed the entire project skeleton to LLMs, providing as much information as possible, \ie, the \textit{\textbf{Maximum Context}} setting.
Note that due to the limitation of long input contexts, it is possible that an LLM fails to digest the entire project skeleton. {Each model has a different context window size, and there is no fixed ratio between code length and token length after tokenization. 
We truncate the first 8192 characters ($\approx$ 3000 tokens) for all studied LLMs.
Then, the smallest maximum window length among the studied LLMs is set to be 8192 tokens, which is larger than the 2,048 tokens set in existing works~\cite{du2023classeval, codereval4python}, ensuring ample space is reserved for output. With this context size, 53.3\% of the contexts are truncated, and the truncated characters account for 42.9\% of the total characters.} 
Opposite to the maximum context, it is natural to use only the class to be completed, \ie, \textbf{\textit{Minimum Context}}.
The advantage of this context setting favors the input token limits, while the disadvantage is the lack of necessary dependencies for synthesis. Take \texttt{ClassA} in Figure~\ref{fig:generate} as an example. In \texttt{ClassA}, a private member \texttt{cb} is declared as an instance of the class \texttt{ClassB}. The Minimum context does not include the code of \texttt{ClassB} into context, which may pose challenges to LLMs to complete methods in \texttt{ClassA}.
Finally, inspired by recent works~\cite{repo-rag-dataflow, Shrivastava23} that use related contexts (\eg, the invoked methods) to strike a balance between rich information and limited input tokens, we take the third context setting, \ie, \textit{\textbf{Selected Context}} into consideration. 
{Specifically, we took advantage of jdeps~\footnote{https://docs.oracle.com/en/java/javase/11/tools/jdeps.html}, which is a command-line tool in the JDK that analyzes Java class files to report on package-level and class-level dependencies, to extract the selected context automatically. }
In addition, to minimize the input tokens while maintaining the context as informative as possible, we \textbf{\textit{only include the method signatures}} in the related class, excluding the method body. For example, as shown in Figure~\ref{fig:generate} (\ding{184} Selected Context), to generate \texttt{func1} in \texttt{ClassA}, the methods in the related \texttt{ClassB} are simplified into signatures.

\subsubsection{\textbf{Synthesis Strategy Design}}\label{sec:syn}
The order of synthesizing methods in each class may matter, so we consider three synthesis strategies following the practice of a prior work~\cite{du2023classeval} and consider two more variants. As shown in the right part in Figure~\ref{fig:generate}, strategies are explained as follows:

\begin{itemize}
    \item \textbf{Independent Synthesis}: each method is synthesized independently without being affected by other generated methods.
    \item \textbf{Holistic Synthesis}: all the methods in a class are synthesized in one pass by LLMs.
    \item \textbf{Incremental Synthesis}: methods in a class are generated one by one in a specific order. Different from prior work~\cite{du2023classeval}, in addition to considering \textit{\textbf{sequential}} the order to synthesize methods, we also consider the \textbf{\textit{reverse}} and \textbf{\textit{random}} orders. 
\end{itemize}

Though according to existing work~\cite{du2023classeval}, these synthesis strategies affect open- and close-sourced LLMs differently, our \name differs in their benchmark, \ie, {\textit{ClassEval}}, in \textbf{\textit{scale}} (class-level vs. project-level) and \textbf{\textit{programming languages}} (Python vs. Java), so similar experiments are still worth exploring on \name. Moreover, we further consider different orders in incremental synthesis, which may yield deeper insights. 

\subsection{Evaluation Design}
Once the entire project is completed by LLMs, we consider two granularities (see Section~\ref{sec:granularity}) to evaluate synthesized projects using three progressive metrics (see Section~\ref{sec:metrics}). 

\subsubsection{\textbf{Evaluation Granularity}}\label{sec:granularity}
Figure~\ref{fig:evaluate} illustrates an example project \texttt{P} with three classes (A, B, and C; canonical ground-truth solutions are included) and three tests (M, N, and P). The central part shows the following two granularities:

\begin{itemize}
    \item \textbf{Class-wise}: To isolate the generated certain class from the others, \textit{class-wise} granularity uses a generated class to replace the canonical solution's counterpart class at a time.
    For example, consider the first column under \textbf{I. Class-wise} setting in Figure~\ref{fig:evaluate}, \texttt{Class A} is generated by LLMs, while {Class B} and \texttt{Class C} are canonical. In other words, \colorbox{greenbg}{Class A (\textbf{Generated})}, \colorbox{greenbg}{Class B (\textit{Canonical})}, and \colorbox{greenbg}{Class C (\textit{Canonical})} form a complete project \textit{P'$_{A}$}. Similarly, \textit{P'$_{B}$} and \textit{P'$_{C}$} are constructed by replacing \texttt{Class B} and \texttt{Class C}, respectively. 
    Then, these projects, each with only one generated class, are evaluated using different metrics (Section~\ref{sec:metrics}), and the average is taken as the result at this granularity.

    \item \textbf{Test-wise}: Similarly, this granularity iterates all the test cases in the test suites and takes the average result. 
    For each test case, we replace the classes relating to the test case while keeping other classes in the canonical solution unchanged. For example, as shown in \textbf{II. Test-wise} setting of Figure~\ref{fig:evaluate}, consider test \texttt{N} which relates to \texttt{Class A} and \texttt{Class C}, we replace these two generated classes while keeping Class B with ground-truth. After enumerating all test cases in the test suite, we evaluate all generated projects using different metrics (Section~\ref{sec:metrics}) and take the average as the result at this granularity.
\end{itemize}

\noindent
We adopt these finer granularities to \textbf{\textit{capture the nuanced difference in performance}}.
Otherwise, the successful generation of some classes could be shadowed by failures in other classes when evaluating at a large granularity.
For example, a \textbf{\textit{project-wise}} evaluation requires the entire generated project to be completed/compiled and pass the entire test suite.
In contrast, class-wise granularity examines one class at a time, allowing for \textbf{\textit{the isolated assessment of each class}}. Section~\ref{sec:rq1} confirmed the effectiveness of such design.

\vspace{-5pt}
\subsubsection{\textbf{Evaluation Metrics}}\label{sec:metrics}
We evaluate the generation code using three progressive metrics: completion/compilation/test case pass rate. 
All are based on the widely used Pass@$k$ metric~\cite{humaneval}:
\begin{equation}
    Completion/Compilation/Pass@k = \mathbb{E} \left [ 1-\binom{n-c}{k} / \binom{n}{k}\right ]
\end{equation}
\noindent where Pass@$k$, as defined in prior work~\cite{humaneval}, is the expectation of passing all the tests of a problem at least once within $k$ attempts. 
For each problem, $n$ solutions are sampled from an LLM, and $c$ of $n$ solutions are correct. 
The larger $n$ is, the more accurate Pass@$k$ is.
Considering the cost and time, we set $n$ to 5 and $k$ to 1 and 5, following the previous study\cite{du2023classeval}.

Similarly, we introduce Completion@k and Compilation@k.
\textbf{\textit{Completion@$k$}} represents the rate of the designated \texttt{TODO} comments being completed in the generated codes.
\textbf{\textit{Compilation@$k$}} represents the rate of the modified projects (replacing parts of the canonical solution at different granularities) being completed and successfully compiled. 
\textbf{\textit{Pass@$k$}} represents the rate of the modified projects being completed, compiled, and passing the test cases related to a specific class (class-wise granularity) or a specific test case with corresponding modified classes (test-wise granularity).

\subsection{\textbf{Prompt Design}}
The prompt used for LLMs generation is shown in Listing~\ref{listing:prompt}. Following the common practice of prompting LLMs~\cite{wizardcoder,du2023classeval}, the prompt template consists of two parts: a \textit{\textbf{system message}} to initialize the model, and an \textbf{\textit{instruction}} to state the purpose of the task.
In Listing~\ref{listing:prompt}, \textcolor{black}{\$\{$\cdot$\}\$} is placeholder: \textcolor{black}{\$\{context\}\$} denotes the context (\eg, three context settings introduced in Section~\ref{sec:context}), and \textcolor{black}{\$\{class\}\$} denotes the class to be completed.

\subsection{\textbf{Studied Large Language Models}}\label{subsec:llms}
The studied LLMs are listed in Table~\ref{tab:models}.
We selected five state-of-the-art LLMs that have been widely explored in code generation tasks.
We focused on recent LLMs (\ie, released after 2022 as the settings in~\cite{du2023classeval}) with more than 6B parameters to achieve sufficient efficacy.
We considered the instruction version of LLMs because we need to utilize the instruction-following ability.
In particular, we selected WizarCoder~\cite{wizardcoder} because it performs better than its base model, StarCoder~\cite{starcoderbase}, in multiple coding tasks.
We chose Phind~\cite{phind} over CodeLlama~\cite{codellama} for the same reason.
We also chose two versions of DeepSeek because they are ranked at the top of the leaderboard. Finally, we include ChatGPT-3.5 because of its popularity and efficacy.
The model size of studied LLMs was at most 34B, limited by our computational resources.

\begin{table}[t!]
\centering
\caption{Studied Large Language Models}
\label{tab:models}
\renewcommand\arraystretch{1.0}
    \resizebox{1.0\linewidth}{!}{
\begin{tabular}{l|l|l|r}
\toprule
\multicolumn{1}{l|}{\textbf{Base Model}} & \multicolumn{1}{c|}{\textbf{Model}} & \multicolumn{1}{c|}{\textbf{Size}} & \multicolumn{1}{c}{\textbf{Time}} \\
\midrule
StarCoder~\cite{li2023starcoder} & WizardCoder-15B-V1.0~\cite{luo2023wizardcoder} & 15B & June, 2023 \\
DeepSeek~\cite{deepseek-llm} & deepseek-coder-6.7b-instruct~\cite{Deepseek6} & 6.7B & Sep, 2023  \\
DeepSeek~\cite{deepseek-llm} & deepseek-coder-33b-instruct~\cite{Deepseek33} & 33B & Nov, 2023  \\
CodeLlama~\cite{codellama} & Phind-CodeLlama-34B-v2~\cite{phindcodellama34bv2} & 34B & Aug, 2023   \\
-- & gpt-3.5-turbo-1106 & -- & Nov, 2022  \\
\bottomrule
\end{tabular}
\vspace{-15pt}
}
\end{table}

\definecolor{stubbg}{HTML}{FCF3D5}
\begin{figure}[h!]
\begin{lstlisting}[
		language=python,
        % basicstyle=\linespread{0.8}\tiny,
        % baselinestretch=1.0,
		% morekeywords={var},
            aboveskip=-4pt,
            belowskip=-15pt,
		caption={\textbf{Prompt Template used in the Experiment}},
		label={listing:prompt},
		escapechar=|,
		% linebackgroundcolor = {\ifnum \value{lstnumber} > 2 \ifnum \value{lstnumber} < 10 \color{stubbg} \fi \fi},
		numbers=left
	]
|\textbf{\textcolor{teal}{\#\# System Message}}|
You are a helpful Java programmer who writes the project based 
on the following context. Java |is| a high-level, |class|-based, 
|object|-oriented programming language designed to have as few 
implementation dependencies as possible.

|\textbf{\textcolor{teal}{\#\# Instruction}}|
```java
|\textcolor{purple}{\textbf{\$\{context\}}}|
```

Complete the code |and| give the completed |class|
```java
|\textcolor{purple}{\textbf{\$\{class\}}}|
```
\end{lstlisting}
\end{figure}

\section{Evaluation}\label{sec:eval}

We used nucleus sampling~\cite{nuclear} in line with recent works~\cite{du2023classeval,yu2023codereval,ouyang2023llm,cao2024concerned}, where five solution samples were randomly generated with a temperature of 0.2~\cite{humaneval}.
The experiments were conducted on a server with two NVIDIA RTX 6000 Ada GPUs, each with 48GB of graphic memory.

The research questions (RQs) were designed as follows:
\begin{itemize}
    \item \textbf{RQ1. Overall Performance. } We first showed the overall performance of the studied LLMs on \name. We used the \textit{selected context} setting and exercised three synthesis strategies (Section~\ref{sec:syn}) to generate the entire project. The comprehensive results were displayed wtih three metrics at two granularities.
    \item \textbf{RQ2. Context Selection.} The context is an important factor in LLMs' performance, so we iterated three context settings (Section~\ref{sec:context}) and observed the corresponding impacts.
    \item \textbf{RQ3. Incremental Strategies. } To synthesize methods in one class, we explored whether the order of synthesizing methods in the class matters. 
    \item \textbf{RQ4. Bad Case Analysis. } {We analyzed five bad cases that failed to compile or pass the test cases due to various issues and identified the incapabilities of LLMs in Java code generation.}
\end{itemize}

\subsection{RQ1: Overall Performance}\label{sec:rq1}

\begin{table*}[th]
\centering
\caption{RQ1 -- Overall Results on \name 
}
\label{tab:rq1}
\renewcommand\arraystretch{1.2}
    \resizebox{1.0\textwidth}{!}{
\begin{tabular}{ll|llll||llll|llll||llll|llll}
\toprule
 &  & \multicolumn{4}{c||}{} & \multicolumn{8}{c||}{\textbf{Compilation@1 (\%)}} & \multicolumn{8}{c}{\textbf{Pass@1} (\%)} \\
 \cline{3-22}
 &  & \multicolumn{4}{c||}{\multirow{-3}{*}{\textbf{Completion@1 (\%)}}} & \multicolumn{4}{c}{\textbf{Class-wise}} & \multicolumn{4}{c||}{\textbf{Test-wise}} & \multicolumn{4}{c}{\textbf{Class-wise}} & \multicolumn{4}{c}{\textbf{Test-wise}}   \\ 
\multicolumn{1}{c}{\textbf{Strategy}} & \multicolumn{1}{c|}{\textbf{Model}} & \multicolumn{1}{c}{P1} & \multicolumn{1}{c}{P2} & \multicolumn{1}{c}{P3} & \multicolumn{1}{c||}{P4} & \multicolumn{1}{c}{P1} & \multicolumn{1}{c}{P2} & \multicolumn{1}{c}{P3} & \multicolumn{1}{c}{P4} & \multicolumn{1}{c}{P1} & \multicolumn{1}{c}{P2} & \multicolumn{1}{c}{P3} & \multicolumn{1}{c||}{P4} & \multicolumn{1}{c}{P1} & \multicolumn{1}{c}{P2} & \multicolumn{1}{c}{P3} & \multicolumn{1}{c}{P4} & \multicolumn{1}{c}{P1} & \multicolumn{1}{c}{P2} & \multicolumn{1}{c}{P3} & \multicolumn{1}{c}{P4} \\
 \midrule
Holistic                     & WizardCoder-15B-V1.0                        & \cellcolor[HTML]{89BDA7}{\color[HTML]{3B3B3B} 98.00}  & \cellcolor[HTML]{ABD0C0}{\color[HTML]{3B3B3B} 70.00} & \cellcolor[HTML]{8DBFAB}{\color[HTML]{3B3B3B} 94.29}  & \cellcolor[HTML]{9FC9B7}{\color[HTML]{3B3B3B} 80.00}  & \cellcolor[HTML]{B2D4C6}{\color[HTML]{3B3B3B} 64.00} & \cellcolor[HTML]{B8D7CA}{\color[HTML]{3B3B3B} 58.57} & \cellcolor[HTML]{A9CFBF}{\color[HTML]{3B3B3B} 71.43} & \cellcolor[HTML]{D2E6DD}{\color[HTML]{3B3B3B} 37.14} & \cellcolor[HTML]{EAF3EF}{\color[HTML]{3B3B3B} 17.50} & \cellcolor[HTML]{FFFFFF}{\color[HTML]{3B3B3B} 0.00}  & \cellcolor[HTML]{D8E9E2}{\color[HTML]{3B3B3B} 31.76} & \cellcolor[HTML]{F8FBF9}{\color[HTML]{3B3B3B} 5.71}  & \cellcolor[HTML]{B2D4C6}{\color[HTML]{3B3B3B} 63.99} & \cellcolor[HTML]{BAD8CC}{\color[HTML]{3B3B3B} 56.87} & \cellcolor[HTML]{AACFC0}{\color[HTML]{3B3B3B} 70.19} & \cellcolor[HTML]{D3E6DE}{\color[HTML]{3B3B3B} 36.14} & \cellcolor[HTML]{EBF3F0}{\color[HTML]{3B3B3B} 16.51} & {\color[HTML]{3B3B3B} 0.00}                         & \cellcolor[HTML]{D9E9E3}{\color[HTML]{3B3B3B} 31.40} & \cellcolor[HTML]{F8FBFA}{\color[HTML]{3B3B3B} 5.03}  \\
Holistic                     & deepseek-coder-6.7b-instruct                & \cellcolor[HTML]{9CC8B6}{\color[HTML]{3B3B3B} 82.00}  & \cellcolor[HTML]{93C2AE}{\color[HTML]{3B3B3B} 90.00} & \cellcolor[HTML]{87BCA6}{\color[HTML]{3B3B3B} 100.00} & \cellcolor[HTML]{87BCA6}{\color[HTML]{3B3B3B} 100.00} & \cellcolor[HTML]{B2D4C6}{\color[HTML]{3B3B3B} 64.00} & \cellcolor[HTML]{A9CFBF}{\color[HTML]{3B3B3B} 71.43} & \cellcolor[HTML]{9BC7B5}{\color[HTML]{3B3B3B} 82.86} & \cellcolor[HTML]{A9CFBF}{\color[HTML]{3B3B3B} 71.43} & \cellcolor[HTML]{EDF4F1}{\color[HTML]{3B3B3B} 15.00} & \cellcolor[HTML]{FFFFFF}{\color[HTML]{3B3B3B} 0.00}  & \cellcolor[HTML]{AACFC0}{\color[HTML]{3B3B3B} 70.59} & \cellcolor[HTML]{BDDACE}{\color[HTML]{3B3B3B} 54.29} & \cellcolor[HTML]{B3D5C7}{\color[HTML]{3B3B3B} 62.67} & \cellcolor[HTML]{AACFBF}{\color[HTML]{3B3B3B} 70.83} & \cellcolor[HTML]{9BC7B5}{\color[HTML]{3B3B3B} 82.55} & \cellcolor[HTML]{B4D5C7}{\color[HTML]{3B3B3B} 62.21} & \cellcolor[HTML]{EDF5F2}{\color[HTML]{3B3B3B} 14.29} & {\color[HTML]{3B3B3B} 0.00}                         & \cellcolor[HTML]{AACFC0}{\color[HTML]{3B3B3B} 70.59} & \cellcolor[HTML]{F2F7F5}{\color[HTML]{3B3B3B} 10.55} \\
Holistic                     & deepseek-coder-33b-instruct                 & \cellcolor[HTML]{87BCA6}{\color[HTML]{3B3B3B} 100.00} & \cellcolor[HTML]{8CBEA9}{\color[HTML]{3B3B3B} 95.71} & \cellcolor[HTML]{87BCA6}{\color[HTML]{3B3B3B} 100.00} & \cellcolor[HTML]{87BCA6}{\color[HTML]{3B3B3B} 100.00} & \cellcolor[HTML]{ADD1C2}{\color[HTML]{3B3B3B} 68.00} & \cellcolor[HTML]{9DC8B6}{\color[HTML]{3B3B3B} 81.43} & \cellcolor[HTML]{A2CBBA}{\color[HTML]{3B3B3B} 77.14} & \cellcolor[HTML]{98C5B2}{\color[HTML]{3B3B3B} 85.71} & \cellcolor[HTML]{B7D6C9}{\color[HTML]{3B3B3B} 60.00} & \cellcolor[HTML]{E9F2EE}{\color[HTML]{3B3B3B} 18.18} & \cellcolor[HTML]{DBEBE4}{\color[HTML]{3B3B3B} 29.41} & \cellcolor[HTML]{A9CFBF}{\color[HTML]{3B3B3B} 71.43} & \cellcolor[HTML]{ADD1C2}{\color[HTML]{3B3B3B} 68.00} & \cellcolor[HTML]{9EC9B7}{\color[HTML]{3B3B3B} 80.49} & \cellcolor[HTML]{A2CBBA}{\color[HTML]{3B3B3B} 76.78} & \cellcolor[HTML]{9CC8B5}{\color[HTML]{3B3B3B} 82.04} & \cellcolor[HTML]{B7D6C9}{\color[HTML]{3B3B3B} 60.00} & \cellcolor[HTML]{F4F9F7}{\color[HTML]{3B3B3B} 8.74} & \cellcolor[HTML]{DBEBE4}{\color[HTML]{3B3B3B} 29.41} & \cellcolor[HTML]{D5E7E0}{\color[HTML]{3B3B3B} 34.34} \\
Holistic                     & Phind-CodeLlama-34B-v2                      & \cellcolor[HTML]{8BBEA9}{\color[HTML]{3B3B3B} 96.00}  & \cellcolor[HTML]{94C3B0}{\color[HTML]{3B3B3B} 88.57} & \cellcolor[HTML]{91C1AD}{\color[HTML]{3B3B3B} 91.43}  & \cellcolor[HTML]{9FC9B7}{\color[HTML]{3B3B3B} 80.00}  & \cellcolor[HTML]{97C5B2}{\color[HTML]{3B3B3B} 86.00} & \cellcolor[HTML]{ACD1C1}{\color[HTML]{3B3B3B} 68.57} & \cellcolor[HTML]{9FC9B7}{\color[HTML]{3B3B3B} 80.00} & \cellcolor[HTML]{BAD8CC}{\color[HTML]{3B3B3B} 57.14} & \cellcolor[HTML]{9FC9B7}{\color[HTML]{3B3B3B} 80.00} & \cellcolor[HTML]{FFFFFF}{\color[HTML]{3B3B3B} 0.00}  & \cellcolor[HTML]{B4D5C7}{\color[HTML]{3B3B3B} 62.35} & \cellcolor[HTML]{E7F1ED}{\color[HTML]{3B3B3B} 20.00} & \cellcolor[HTML]{98C5B2}{\color[HTML]{3B3B3B} 85.81} & \cellcolor[HTML]{ADD1C2}{\color[HTML]{3B3B3B} 67.54} & \cellcolor[HTML]{9FC9B8}{\color[HTML]{3B3B3B} 79.66} & \cellcolor[HTML]{BDDACE}{\color[HTML]{3B3B3B} 54.58} & \cellcolor[HTML]{B9D8CB}{\color[HTML]{3B3B3B} 58.06} & {\color[HTML]{3B3B3B} 0.00}                         & \cellcolor[HTML]{B4D5C7}{\color[HTML]{3B3B3B} 62.32} & \cellcolor[HTML]{F4F9F7}{\color[HTML]{3B3B3B} 8.42}  \\
Holistic                     & gpt-3.5-turbo-1106                          & \cellcolor[HTML]{93C2AE}{\color[HTML]{3B3B3B} 90.00}  & \cellcolor[HTML]{99C6B3}{\color[HTML]{3B3B3B} 84.29} & \cellcolor[HTML]{87BCA6}{\color[HTML]{3B3B3B} 100.00} & \cellcolor[HTML]{8DBFAB}{\color[HTML]{3B3B3B} 94.29}  & \cellcolor[HTML]{97C5B2}{\color[HTML]{3B3B3B} 86.00} & \cellcolor[HTML]{A2CBBA}{\color[HTML]{3B3B3B} 77.14} & \cellcolor[HTML]{99C6B3}{\color[HTML]{3B3B3B} 84.29} & \cellcolor[HTML]{A5CDBC}{\color[HTML]{3B3B3B} 74.29} & \cellcolor[HTML]{A5CCBC}{\color[HTML]{3B3B3B} 75.00} & \cellcolor[HTML]{FFFFFF}{\color[HTML]{3B3B3B} 0.00}  & \cellcolor[HTML]{AACFC0}{\color[HTML]{3B3B3B} 70.59} & \cellcolor[HTML]{EAF3EF}{\color[HTML]{3B3B3B} 17.14} & \cellcolor[HTML]{97C5B2}{\color[HTML]{3B3B3B} 85.87} & \cellcolor[HTML]{A3CCBB}{\color[HTML]{3B3B3B} 75.98} & \cellcolor[HTML]{9AC6B4}{\color[HTML]{3B3B3B} 83.90} & \cellcolor[HTML]{A8CEBE}{\color[HTML]{3B3B3B} 72.24} & \cellcolor[HTML]{B4D5C7}{\color[HTML]{3B3B3B} 62.29} & {\color[HTML]{3B3B3B} 0.00}                         & \cellcolor[HTML]{AACFC0}{\color[HTML]{3B3B3B} 70.59} & \cellcolor[HTML]{F8FBFA}{\color[HTML]{3B3B3B} 5.40}  \\
\hline
 \multicolumn{2}{r|}{\textbf{Average} (\textbf{\textit{Holistic}})} &\multicolumn{4}{c||}{\textbf{91.73}} & \multicolumn{4}{c|}{\textbf{72.33}} & \multicolumn{4}{c||}{\textbf{34.95}} & \multicolumn{4}{c|}{\textbf{70.92}} & \multicolumn{4}{c}{\textbf{27.40}} \\
\hline 

Independent                  & WizardCoder-15B-V1.0                        & \cellcolor[HTML]{A6CDBD}{\color[HTML]{3B3B3B} 74.00}  & \cellcolor[HTML]{A7CEBE}{\color[HTML]{3B3B3B} 72.86} & \cellcolor[HTML]{A9CFBF}{\color[HTML]{3B3B3B} 71.43}  & \cellcolor[HTML]{D9E9E3}{\color[HTML]{3B3B3B} 31.43}  & \cellcolor[HTML]{CAE1D7}{\color[HTML]{3B3B3B} 44.00} & \cellcolor[HTML]{B3D4C7}{\color[HTML]{3B3B3B} 62.86} & \cellcolor[HTML]{C6DFD5}{\color[HTML]{3B3B3B} 47.14} & \cellcolor[HTML]{EAF3EF}{\color[HTML]{3B3B3B} 17.14} & \cellcolor[HTML]{FFFFFF}{\color[HTML]{3B3B3B} 0.00}  & \cellcolor[HTML]{FFFFFF}{\color[HTML]{3B3B3B} 0.00}  & \cellcolor[HTML]{FFFFFF}{\color[HTML]{3B3B3B} 0.00}  & \cellcolor[HTML]{FFFFFF}{\color[HTML]{3B3B3B} 0.00}  & \cellcolor[HTML]{CAE1D7}{\color[HTML]{3B3B3B} 44.00} & \cellcolor[HTML]{B5D6C8}{\color[HTML]{3B3B3B} 61.17} & \cellcolor[HTML]{C6DFD5}{\color[HTML]{3B3B3B} 46.92} & \cellcolor[HTML]{EBF3F0}{\color[HTML]{3B3B3B} 16.62} & \cellcolor[HTML]{FFFFFF}{\color[HTML]{3B3B3B} 0.00}  & {\color[HTML]{3B3B3B} 0.00}                         & \cellcolor[HTML]{FFFFFF}{\color[HTML]{3B3B3B} 0.00}  & \cellcolor[HTML]{FFFFFF}{\color[HTML]{3B3B3B} 0.00}  \\
Independent                  & deepseek-coder-6.7b-instruct                & \cellcolor[HTML]{9FC9B7}{\color[HTML]{3B3B3B} 80.00}  & \cellcolor[HTML]{93C2AE}{\color[HTML]{3B3B3B} 90.00} & \cellcolor[HTML]{8ABDA8}{\color[HTML]{3B3B3B} 97.14}  & \cellcolor[HTML]{98C5B2}{\color[HTML]{3B3B3B} 85.71}  & \cellcolor[HTML]{ABD0C0}{\color[HTML]{3B3B3B} 70.00} & \cellcolor[HTML]{AED2C3}{\color[HTML]{3B3B3B} 67.14} & \cellcolor[HTML]{A0CAB9}{\color[HTML]{3B3B3B} 78.57} & \cellcolor[HTML]{B3D4C7}{\color[HTML]{3B3B3B} 62.86} & \cellcolor[HTML]{D5E7DF}{\color[HTML]{3B3B3B} 35.00} & \cellcolor[HTML]{FFFFFF}{\color[HTML]{3B3B3B} 0.00}  & \cellcolor[HTML]{AACFC0}{\color[HTML]{3B3B3B} 70.59} & \cellcolor[HTML]{CBE2D8}{\color[HTML]{3B3B3B} 42.86} & \cellcolor[HTML]{ABD0C1}{\color[HTML]{3B3B3B} 69.29} & \cellcolor[HTML]{AFD2C3}{\color[HTML]{3B3B3B} 66.40} & \cellcolor[HTML]{A0CAB9}{\color[HTML]{3B3B3B} 78.57} & \cellcolor[HTML]{BCD9CD}{\color[HTML]{3B3B3B} 55.33} & \cellcolor[HTML]{D9E9E2}{\color[HTML]{3B3B3B} 31.55} & {\color[HTML]{3B3B3B} 0.00}                         & \cellcolor[HTML]{AACFC0}{\color[HTML]{3B3B3B} 70.59} & \cellcolor[HTML]{EFF6F3}{\color[HTML]{3B3B3B} 12.75} \\
Independent                  & deepseek-coder-33b-instruct                 & \cellcolor[HTML]{A3CCBB}{\color[HTML]{3B3B3B} 76.00}  & \cellcolor[HTML]{A0CAB9}{\color[HTML]{3B3B3B} 78.57} & \cellcolor[HTML]{8FC0AC}{\color[HTML]{3B3B3B} 92.86}  & \cellcolor[HTML]{87BCA6}{\color[HTML]{3B3B3B} 100.00} & \cellcolor[HTML]{AFD2C4}{\color[HTML]{3B3B3B} 66.00} & \cellcolor[HTML]{A4CCBB}{\color[HTML]{3B3B3B} 75.71} & \cellcolor[HTML]{A5CDBC}{\color[HTML]{3B3B3B} 74.29} & \cellcolor[HTML]{A5CDBC}{\color[HTML]{3B3B3B} 74.29} & \cellcolor[HTML]{B4D5C7}{\color[HTML]{3B3B3B} 62.50} & \cellcolor[HTML]{FFFFFF}{\color[HTML]{3B3B3B} 0.00}  & \cellcolor[HTML]{D1E5DD}{\color[HTML]{3B3B3B} 37.65} & \cellcolor[HTML]{B7D6C9}{\color[HTML]{3B3B3B} 60.00} & \cellcolor[HTML]{AFD2C4}{\color[HTML]{3B3B3B} 66.00} & \cellcolor[HTML]{A6CDBD}{\color[HTML]{3B3B3B} 73.50} & \cellcolor[HTML]{A5CDBC}{\color[HTML]{3B3B3B} 74.26} & \cellcolor[HTML]{ABD0C1}{\color[HTML]{3B3B3B} 69.45} & \cellcolor[HTML]{BAD8CC}{\color[HTML]{3B3B3B} 56.82} & {\color[HTML]{3B3B3B} 0.00}                         & \cellcolor[HTML]{D1E5DD}{\color[HTML]{3B3B3B} 37.54} & \cellcolor[HTML]{E5F0EC}{\color[HTML]{3B3B3B} 21.07} \\
Independent                  & Phind-CodeLlama-34B-v2                      & \cellcolor[HTML]{A6CDBD}{\color[HTML]{3B3B3B} 74.00}  & \cellcolor[HTML]{96C4B1}{\color[HTML]{3B3B3B} 87.14} & \cellcolor[HTML]{ACD1C1}{\color[HTML]{3B3B3B} 68.57}  & \cellcolor[HTML]{A9CFBF}{\color[HTML]{3B3B3B} 71.43}  & \cellcolor[HTML]{B2D4C6}{\color[HTML]{3B3B3B} 64.00} & \cellcolor[HTML]{ACD1C1}{\color[HTML]{3B3B3B} 68.57} & \cellcolor[HTML]{B5D5C8}{\color[HTML]{3B3B3B} 61.43} & \cellcolor[HTML]{C4DED3}{\color[HTML]{3B3B3B} 48.57} & \cellcolor[HTML]{F0F6F3}{\color[HTML]{3B3B3B} 12.50} & \cellcolor[HTML]{FFFFFF}{\color[HTML]{3B3B3B} 0.00}  & \cellcolor[HTML]{DFEDE7}{\color[HTML]{3B3B3B} 25.88} & \cellcolor[HTML]{F1F7F4}{\color[HTML]{3B3B3B} 11.43} & \cellcolor[HTML]{B2D4C6}{\color[HTML]{3B3B3B} 63.89} & \cellcolor[HTML]{AED1C3}{\color[HTML]{3B3B3B} 67.38} & \cellcolor[HTML]{B5D5C8}{\color[HTML]{3B3B3B} 61.30} & \cellcolor[HTML]{C6DFD5}{\color[HTML]{3B3B3B} 46.96} & \cellcolor[HTML]{F0F6F3}{\color[HTML]{3B3B3B} 12.50} & {\color[HTML]{3B3B3B} 0.00}                         & \cellcolor[HTML]{DFEDE7}{\color[HTML]{3B3B3B} 25.88} & \cellcolor[HTML]{FBFDFC}{\color[HTML]{3B3B3B} 2.71}  \\
Independent                  & gpt-3.5-turbo-1106                          & \cellcolor[HTML]{93C2AE}{\color[HTML]{3B3B3B} 90.00}  & \cellcolor[HTML]{94C3B0}{\color[HTML]{3B3B3B} 88.57} & \cellcolor[HTML]{8FC0AC}{\color[HTML]{3B3B3B} 92.86}  & \cellcolor[HTML]{A9CFBF}{\color[HTML]{3B3B3B} 71.43}  & \cellcolor[HTML]{A1CAB9}{\color[HTML]{3B3B3B} 78.00} & \cellcolor[HTML]{9DC8B6}{\color[HTML]{3B3B3B} 81.43} & \cellcolor[HTML]{A7CEBE}{\color[HTML]{3B3B3B} 72.86} & \cellcolor[HTML]{CBE2D8}{\color[HTML]{3B3B3B} 42.86} & \cellcolor[HTML]{F0F6F3}{\color[HTML]{3B3B3B} 12.50} & \cellcolor[HTML]{FFFFFF}{\color[HTML]{3B3B3B} 0.00}  & \cellcolor[HTML]{B1D3C5}{\color[HTML]{3B3B3B} 64.71} & \cellcolor[HTML]{FFFFFF}{\color[HTML]{3B3B3B} 0.00}  & \cellcolor[HTML]{A1CAB9}{\color[HTML]{3B3B3B} 77.80} & \cellcolor[HTML]{9EC8B7}{\color[HTML]{3B3B3B} 80.80} & \cellcolor[HTML]{A7CEBE}{\color[HTML]{3B3B3B} 72.86} & \cellcolor[HTML]{CBE2D9}{\color[HTML]{3B3B3B} 42.64} & \cellcolor[HTML]{F0F6F3}{\color[HTML]{3B3B3B} 12.50} & {\color[HTML]{3B3B3B} 0.00}                         & \cellcolor[HTML]{B1D3C5}{\color[HTML]{3B3B3B} 64.71} & \cellcolor[HTML]{FFFFFF}{\color[HTML]{3B3B3B} 0.00}  \\
\hline
\multicolumn{2}{r|}{\textbf{Average} (\textit{\textbf{Independent}})} &\multicolumn{4}{c||}{{79.70}} & \multicolumn{4}{c|}{{62.89}} & \multicolumn{4}{c||}{{21.78}} & \multicolumn{4}{c|}{{61.76}} & \multicolumn{4}{c}{{17.43}} \\
\hline

Incremental                  & WizardCoder-15B-V1.0                        & \cellcolor[HTML]{ABD0C0}{\color[HTML]{3B3B3B} 70.00}  & \cellcolor[HTML]{A5CDBC}{\color[HTML]{3B3B3B} 74.29} & \cellcolor[HTML]{C4DED3}{\color[HTML]{3B3B3B} 48.57}  & \cellcolor[HTML]{DCEBE5}{\color[HTML]{3B3B3B} 28.57}  & \cellcolor[HTML]{C7E0D6}{\color[HTML]{3B3B3B} 46.00} & \cellcolor[HTML]{C1DCD1}{\color[HTML]{3B3B3B} 51.43} & \cellcolor[HTML]{D4E7DF}{\color[HTML]{3B3B3B} 35.71} & \cellcolor[HTML]{EAF3EF}{\color[HTML]{3B3B3B} 17.14} & \cellcolor[HTML]{FFFFFF}{\color[HTML]{3B3B3B} 0.00}  & \cellcolor[HTML]{FFFFFF}{\color[HTML]{3B3B3B} 0.00}  & \cellcolor[HTML]{FFFFFF}{\color[HTML]{3B3B3B} 0.00}  & \cellcolor[HTML]{FFFFFF}{\color[HTML]{3B3B3B} 0.00}  & \cellcolor[HTML]{C8E0D6}{\color[HTML]{3B3B3B} 45.31} & \cellcolor[HTML]{C1DCD1}{\color[HTML]{3B3B3B} 50.84} & \cellcolor[HTML]{D4E7DF}{\color[HTML]{3B3B3B} 35.68} & \cellcolor[HTML]{EBF3F0}{\color[HTML]{3B3B3B} 16.62} & \cellcolor[HTML]{FFFFFF}{\color[HTML]{3B3B3B} 0.00}  & {\color[HTML]{3B3B3B} 0.00}                         & \cellcolor[HTML]{FFFFFF}{\color[HTML]{3B3B3B} 0.00}  & \cellcolor[HTML]{FFFFFF}{\color[HTML]{3B3B3B} 0.00}  \\
Incremental                  & deepseek-coder-6.7b-instruct                & \cellcolor[HTML]{A8CEBE}{\color[HTML]{3B3B3B} 72.00}  & \cellcolor[HTML]{93C2AE}{\color[HTML]{3B3B3B} 90.00} & \cellcolor[HTML]{88BCA7}{\color[HTML]{3B3B3B} 98.57}  & \cellcolor[HTML]{91C1AD}{\color[HTML]{3B3B3B} 91.43}  & \cellcolor[HTML]{ADD1C2}{\color[HTML]{3B3B3B} 68.00} & \cellcolor[HTML]{A7CEBE}{\color[HTML]{3B3B3B} 72.86} & \cellcolor[HTML]{9BC7B5}{\color[HTML]{3B3B3B} 82.86} & \cellcolor[HTML]{ACD1C1}{\color[HTML]{3B3B3B} 68.57} & \cellcolor[HTML]{FFFFFF}{\color[HTML]{3B3B3B} 0.00}  & \cellcolor[HTML]{FFFFFF}{\color[HTML]{3B3B3B} 0.00}  & \cellcolor[HTML]{AACFC0}{\color[HTML]{3B3B3B} 70.59} & \cellcolor[HTML]{D5E8E0}{\color[HTML]{3B3B3B} 34.29} & \cellcolor[HTML]{ADD1C2}{\color[HTML]{3B3B3B} 67.51} & \cellcolor[HTML]{A8CEBF}{\color[HTML]{3B3B3B} 71.89} & \cellcolor[HTML]{9CC7B5}{\color[HTML]{3B3B3B} 82.15} & \cellcolor[HTML]{B1D3C5}{\color[HTML]{3B3B3B} 64.30} & \cellcolor[HTML]{FFFFFF}{\color[HTML]{3B3B3B} 0.00}  & {\color[HTML]{3B3B3B} 0.00}                         & \cellcolor[HTML]{AACFC0}{\color[HTML]{3B3B3B} 70.59} & \cellcolor[HTML]{F5F9F7}{\color[HTML]{3B3B3B} 8.12}  \\
Incremental                  & deepseek-coder-33b-instruct                 & \cellcolor[HTML]{ADD1C2}{\color[HTML]{3B3B3B} 68.00}  & \cellcolor[HTML]{A2CBBA}{\color[HTML]{3B3B3B} 77.14} & \cellcolor[HTML]{8FC0AC}{\color[HTML]{3B3B3B} 92.86}  & \cellcolor[HTML]{98C5B2}{\color[HTML]{3B3B3B} 85.71}  & \cellcolor[HTML]{BEDACE}{\color[HTML]{3B3B3B} 54.00} & \cellcolor[HTML]{A5CDBC}{\color[HTML]{3B3B3B} 74.29} & \cellcolor[HTML]{A9CFBF}{\color[HTML]{3B3B3B} 71.43} & \cellcolor[HTML]{BDDACE}{\color[HTML]{3B3B3B} 54.29} & \cellcolor[HTML]{BDDACE}{\color[HTML]{3B3B3B} 55.00} & \cellcolor[HTML]{FFFFFF}{\color[HTML]{3B3B3B} 0.00}  & \cellcolor[HTML]{EEF5F2}{\color[HTML]{3B3B3B} 14.12} & \cellcolor[HTML]{FFFFFF}{\color[HTML]{3B3B3B} 0.00}  & \cellcolor[HTML]{BEDACE}{\color[HTML]{3B3B3B} 54.00} & \cellcolor[HTML]{A6CDBD}{\color[HTML]{3B3B3B} 73.99} & \cellcolor[HTML]{A9CFBF}{\color[HTML]{3B3B3B} 71.38} & \cellcolor[HTML]{C3DDD2}{\color[HTML]{3B3B3B} 49.81} & \cellcolor[HTML]{C4DED3}{\color[HTML]{3B3B3B} 49.08} & {\color[HTML]{3B3B3B} 0.00}                         & \cellcolor[HTML]{EEF5F2}{\color[HTML]{3B3B3B} 14.02} & \cellcolor[HTML]{FFFFFF}{\color[HTML]{3B3B3B} 0.00}  \\
Incremental                  & Phind-CodeLlama-34B-v2                      & \cellcolor[HTML]{A8CEBE}{\color[HTML]{3B3B3B} 72.00}  & \cellcolor[HTML]{96C4B1}{\color[HTML]{3B3B3B} 87.14} & \cellcolor[HTML]{AED2C3}{\color[HTML]{3B3B3B} 67.14}  & \cellcolor[HTML]{BDDACE}{\color[HTML]{3B3B3B} 54.29}  & \cellcolor[HTML]{B4D5C7}{\color[HTML]{3B3B3B} 62.00} & \cellcolor[HTML]{A7CEBE}{\color[HTML]{3B3B3B} 72.86} & \cellcolor[HTML]{B7D6C9}{\color[HTML]{3B3B3B} 60.00} & \cellcolor[HTML]{CBE2D8}{\color[HTML]{3B3B3B} 42.86} & \cellcolor[HTML]{F0F6F3}{\color[HTML]{3B3B3B} 12.50} & \cellcolor[HTML]{FFFFFF}{\color[HTML]{3B3B3B} 0.00}  & \cellcolor[HTML]{E4F0EB}{\color[HTML]{3B3B3B} 22.35} & \cellcolor[HTML]{F8FBF9}{\color[HTML]{3B3B3B} 5.71}  & \cellcolor[HTML]{B4D5C7}{\color[HTML]{3B3B3B} 61.99} & \cellcolor[HTML]{A8CEBE}{\color[HTML]{3B3B3B} 72.05} & \cellcolor[HTML]{B8D7CA}{\color[HTML]{3B3B3B} 59.14} & \cellcolor[HTML]{CDE3DA}{\color[HTML]{3B3B3B} 41.01} & \cellcolor[HTML]{F0F6F3}{\color[HTML]{3B3B3B} 12.50} & {\color[HTML]{3B3B3B} 0.00}                         & \cellcolor[HTML]{E4F0EB}{\color[HTML]{3B3B3B} 21.87} & \cellcolor[HTML]{FBFDFC}{\color[HTML]{3B3B3B} 2.53}  \\
Incremental                  & gpt-3.5-turbo-1106                          & \cellcolor[HTML]{A8CEBE}{\color[HTML]{3B3B3B} 72.00}  & \cellcolor[HTML]{98C5B2}{\color[HTML]{3B3B3B} 85.71} & \cellcolor[HTML]{93C2AE}{\color[HTML]{3B3B3B} 90.00}  & \cellcolor[HTML]{91C1AD}{\color[HTML]{3B3B3B} 91.43}  & \cellcolor[HTML]{B4D5C7}{\color[HTML]{3B3B3B} 62.00} & \cellcolor[HTML]{9FC9B7}{\color[HTML]{3B3B3B} 80.00} & \cellcolor[HTML]{A9CFBF}{\color[HTML]{3B3B3B} 71.43} & \cellcolor[HTML]{ACD1C1}{\color[HTML]{3B3B3B} 68.57} & \cellcolor[HTML]{F3F8F6}{\color[HTML]{3B3B3B} 10.00} & \cellcolor[HTML]{FFFFFF}{\color[HTML]{3B3B3B} 0.00}  & \cellcolor[HTML]{FFFFFF}{\color[HTML]{3B3B3B} 0.00}  & \cellcolor[HTML]{E0EDE8}{\color[HTML]{3B3B3B} 25.71} & \cellcolor[HTML]{B5D5C8}{\color[HTML]{3B3B3B} 61.50} & \cellcolor[HTML]{9FC9B7}{\color[HTML]{3B3B3B} 79.80} & \cellcolor[HTML]{A9CFBF}{\color[HTML]{3B3B3B} 71.43} & \cellcolor[HTML]{B1D3C5}{\color[HTML]{3B3B3B} 64.84} & \cellcolor[HTML]{F3F8F6}{\color[HTML]{3B3B3B} 10.00} & {\color[HTML]{3B3B3B} 0.00}                         & \cellcolor[HTML]{FFFFFF}{\color[HTML]{3B3B3B} 0.00}  & \cellcolor[HTML]{F5F9F7}{\color[HTML]{3B3B3B} 8.10}  \\

\hline
\multicolumn{2}{r|}{\textbf{Average} (\textbf{\textit{Incremental}})} &\multicolumn{4}{c||}{{75.84}} & \multicolumn{4}{c|}{{60.81}} & \multicolumn{4}{c||}{{12.51}} & \multicolumn{4}{c|}{{59.76}} & \multicolumn{4}{c}{{9.84}} \\
\midrule

\multicolumn{2}{r|}{\textbf{Average} } &\multicolumn{4}{c||}{\textbf{82.42}} & \multicolumn{4}{c|}{\textbf{65.34}} & \multicolumn{4}{c||}{\textbf{23.08}} & \multicolumn{4}{c|}{\textbf{64.15}} & \multicolumn{4}{c}{\textbf{18.22}} \\

\bottomrule
\end{tabular}
}
\end{table*}

The overall performance of the studied LLMs on \name was shown in Table~\ref{tab:rq1}.
We first fixed the context setting (\ie, the \textit{selected context})
and exercised three synthesis strategies (\ie, \textit{Holistic}, \textit{Independent}, and \textit{Incremental}) as shown in Figure~\ref{fig:generate}.
Three evaluation metrics (\textit{Completion@1}, \textit{Compilation@1} and \textit{Pass@1}) were three main dimensions in Table~\ref{tab:rq1}.
To better visualize the results, we use darker background colors to indicate larger values.
It is clear that from left to right, the color became lighter, meaning the value was getting smaller as the three metrics got stricter -- Only those completed codes had chance to be compiled (incomplete codes were treated failed when computing Compilation@$k$), and only those compiled codes had chance to be evaluated against test cases (uncompilable codes were treated failed when computing Pass@$k$).

Generally, the average Completion@1 of all LLMs on \name was 82.42\%, with variances when different synthesis strategies were applied. 
Considering a finer granularity (class-wise), the average Compilation@1 and Pass@1 were around 65\% (65.34\% and 64.15\%, respectively).
The best scores were achieved using the holistic strategy with 91.73\% Completion@1, 72.33\% Compilation@1, and 70.92\% Pass@1.
Among all LLMs, DeepSeek-Coder-33b performed the best, followed by gpt-3.5-turbo and DeepSeek-Coder-6.7b. 

\begin{mdframed}[style=MyFrame]
\textbf{Finding 1:} The best average of \textbf{\textit{91.73\%, 72.33\%, and 70.92\%}} Completion@1, Compilation@1, and Pass@1, respectively, could be achieved on \name over the studied LLMs.
The Top-3 performing LLMs were DeepSeek-Coder-33b, gpt-3.5-turbo, and DeepSeek-Coder-6.7b among the studied LLMs.
\end{mdframed}

Comparing the synthesis strategies, \textbf{\textit{holistic}} synthesis was generally better than independent and incremental among all LLMs, and the declines could be significant in some cases.
For example, WizardCoder \textbf{dropped 51.43\%} (80.00\% - 28.57\%) Completion@1,  dropped 20.00\% (37.14\% - 17.14\%) Compilation@1 and 19.52\% (36.14\% - 16.62\%) Pass@1 on P4.
Other LLMs also observed similar drops when switching from holistic to independent or incremental synthesis.
Although there were occasional cases where incremental or independent synthesis brought improvements, those were sporadic events, and the improvement was subtle, \eg, Completion@1 of WizardCoder improved 4.29\% (74.29\% - 70.00\%) on P2.
We also noticed that this observation was different from the observation of the previous work~\cite{du2023classeval} where they observed open-sourced LLMs performed better using independent synthesis than holistic synthesis.
This could be because we used different \textbf{\textit{programming languages}} (Python VS. Java) and \textbf{\textit{code granularities}} (Class-level VS. Project-level).

\begin{mdframed}[style=MyFrame]
\textbf{Finding 2:} Among the three synthesis strategies, \textbf{\textit{holistic synthesis}} yielded a better performance across all LLMs against three metrics at two granularities on \name. 
\end{mdframed}

\textbf{\textit{Necessity of Finer-grained Evaluation Granularity.}} 
LLMs performed similarly on four projects (P1-P4), with P1 and P3 slightly better than P2 and P4. 
In particular, the class-wise scores were always better than test-wise scores, with a gap up to 49.92\% (= 59.76\%-9.84\%).
This was in line with our claim in Section~\ref{sec:granularity}: finer-grained granularity can capture more nuanced successful cases.
We also calculated the Pass@1 under \textbf{\textit{project-wise}} granularity and found that \textbf{\textit{none of the projects can be correctly completed}}, yielding all-0 results under all settings. 

\begin{mdframed}[style=MyFrame]
\textbf{Finding 3: } The \textbf{\textit{finer granularities}} (class-/test-wise) can capture subtle success in performance, enabling \textbf{\textit{more distinguishable}} results. Otherwise, the subtle success can be shadowed by other failures, leading to all-0 results under all settings.
\end{mdframed}

In addition, we increased the size of $k$ from 1 to 5 to investigate the improvements brought by more trials.
The detailed experiment results are omitted due to space limits and can be found on our website (Section~\ref{sec:data}).
Overall, the average \textbf{\textit{Completion/Compilation/Pass@5}}  under holistic strategy across all LLMs are 97.21\%(+5.48\%), 84.43\%(+12.10\%), 84.43\%(+13.51\%) at the \textit{class-wise} granularity; 94.21\%(+5.48\%), 51.23\%(+16.28\%), \textbf{48.24\%(+20.84\%)} at the \textit{test-wise} granularity.
The Pass@5 in \textit{project-wise} granularity is still all-0s under all settings.
\begin{mdframed}[style=MyFrame]
\textbf{Finding 4: }
Increasing $k$ yields further improvement. The best average test-wise Pass@5 in \name is 48.24\%. Compared with 90.93\% achieved by undergraduate students in project-wise evaluation (Section~\ref{sec:human}), LLMs' capability in Java project-level programming still has much room to improve. 
\end{mdframed}
\subsection{RQ2: Impact of Context Selection}\label{sec:rq2}

\begin{table*}[t!]
\centering
\caption{RQ2 -- Performance Variance Over Different Context Settings.}
\label{tab:rq2}
\renewcommand\arraystretch{1.2}
    \resizebox{1.0\textwidth}{!}{
\begin{tabular}{ll|llll||llll||llll||llll||llll}
\toprule
 &  & \multicolumn{4}{c||}{} & \multicolumn{8}{c||}{\textbf{Compilation @ 1 (\%)}} & \multicolumn{8}{c}{\textbf{Pass @ 1} (\%)} \\
 \cline{3-22}
 &  & \multicolumn{4}{c||}{\multirow{-3}{*}{\textbf{Completion @ 1 (\%)}}} & \multicolumn{4}{c}{\textbf{Class-wise}} & \multicolumn{4}{c||}{\textbf{Test-wise}} & \multicolumn{4}{c}{\textbf{Class-wise}} & \multicolumn{4}{c}{\textbf{Test-wise}}   \\ 
\multicolumn{1}{l}{\textbf{Context}} & \multicolumn{1}{c|}{\textbf{Model}} & \multicolumn{1}{c}{P1} & \multicolumn{1}{c}{P2} & \multicolumn{1}{c}{P3} & \multicolumn{1}{c||}{P4} & \multicolumn{1}{c}{P1} & \multicolumn{1}{c}{P2} & \multicolumn{1}{c}{P3} & \multicolumn{1}{c}{P4} & \multicolumn{1}{c}{P1} & \multicolumn{1}{c}{P2} & \multicolumn{1}{c}{P3} & \multicolumn{1}{c||}{P4} & \multicolumn{1}{c}{P1} & \multicolumn{1}{c}{P2} & \multicolumn{1}{c}{P3} & \multicolumn{1}{c}{P4} & \multicolumn{1}{c}{P1} & \multicolumn{1}{c}{P2} & \multicolumn{1}{c}{P3} & \multicolumn{1}{c}{P4} \\
 \midrule
 
Selected & WizardCoder-15B-V1.0 & \cellcolor[HTML]{89BDA7}{\color[HTML]{3B3B3B} 98.00} & \cellcolor[HTML]{ABD0C0}{\color[HTML]{3B3B3B} 70.00} & \cellcolor[HTML]{8DBFAB}{\color[HTML]{3B3B3B} 94.29} & \cellcolor[HTML]{9FC9B7}{\color[HTML]{3B3B3B} 80.00} & \cellcolor[HTML]{B2D4C6}{\color[HTML]{3B3B3B} 64.00} & \cellcolor[HTML]{B8D7CA}{\color[HTML]{3B3B3B} 58.57} & \cellcolor[HTML]{A9CFBF}{\color[HTML]{3B3B3B} 71.43} & \cellcolor[HTML]{D2E6DD}{\color[HTML]{3B3B3B} 37.14} & \cellcolor[HTML]{EAF3EF}{\color[HTML]{3B3B3B} 17.50} & \cellcolor[HTML]{FFFFFF}{\color[HTML]{3B3B3B} 0.00} & \cellcolor[HTML]{D8E9E2}{\color[HTML]{3B3B3B} 31.76} & \cellcolor[HTML]{F8FBF9}{\color[HTML]{3B3B3B} 5.71} & \cellcolor[HTML]{B2D4C6}{\color[HTML]{3B3B3B} 63.99} & \cellcolor[HTML]{BAD8CC}{\color[HTML]{3B3B3B} 56.87} & \cellcolor[HTML]{AACFC0}{\color[HTML]{3B3B3B} 70.19} & \cellcolor[HTML]{D3E6DE}{\color[HTML]{3B3B3B} 36.14} & \cellcolor[HTML]{EBF3F0}{\color[HTML]{3B3B3B} 16.51} & \cellcolor[HTML]{FFFFFF}{\color[HTML]{3B3B3B} 0.00} & \cellcolor[HTML]{D9E9E3}{\color[HTML]{3B3B3B} 31.40} & \cellcolor[HTML]{F8FBFA}{\color[HTML]{3B3B3B} 5.03} \\
Selected & deepseek-coder-6.7b-instruct & \cellcolor[HTML]{9CC8B6}{\color[HTML]{3B3B3B} 82.00} & \cellcolor[HTML]{93C2AE}{\color[HTML]{3B3B3B} 90.00} & \cellcolor[HTML]{87BCA6}{\color[HTML]{3B3B3B} 100.00} & \cellcolor[HTML]{87BCA6}{\color[HTML]{3B3B3B} 100.00} & \cellcolor[HTML]{B2D4C6}{\color[HTML]{3B3B3B} 64.00} & \cellcolor[HTML]{A9CFBF}{\color[HTML]{3B3B3B} 71.43} & \cellcolor[HTML]{9BC7B5}{\color[HTML]{3B3B3B} 82.86} & \cellcolor[HTML]{A9CFBF}{\color[HTML]{3B3B3B} 71.43} & \cellcolor[HTML]{EDF4F1}{\color[HTML]{3B3B3B} 15.00} & \cellcolor[HTML]{FFFFFF}{\color[HTML]{3B3B3B} 0.00} & \cellcolor[HTML]{AACFC0}{\color[HTML]{3B3B3B} 70.59} & \cellcolor[HTML]{BDDACE}{\color[HTML]{3B3B3B} 54.29} & \cellcolor[HTML]{B3D5C7}{\color[HTML]{3B3B3B} 62.67} & \cellcolor[HTML]{AACFBF}{\color[HTML]{3B3B3B} 70.83} & \cellcolor[HTML]{9BC7B5}{\color[HTML]{3B3B3B} 82.55} & \cellcolor[HTML]{B4D5C7}{\color[HTML]{3B3B3B} 62.21} & \cellcolor[HTML]{EDF5F2}{\color[HTML]{3B3B3B} 14.29} & \cellcolor[HTML]{FFFFFF}{\color[HTML]{3B3B3B} 0.00} & \cellcolor[HTML]{AACFC0}{\color[HTML]{3B3B3B} 70.59} & \cellcolor[HTML]{F2F7F5}{\color[HTML]{3B3B3B} 10.55} \\
Selected & deepseek-coder-33b-instruct & \cellcolor[HTML]{87BCA6}{\color[HTML]{3B3B3B} 100.00} & \cellcolor[HTML]{8CBEA9}{\color[HTML]{3B3B3B} 95.71} & \cellcolor[HTML]{87BCA6}{\color[HTML]{3B3B3B} 100.00} & \cellcolor[HTML]{87BCA6}{\color[HTML]{3B3B3B} 100.00} & \cellcolor[HTML]{ADD1C2}{\color[HTML]{3B3B3B} 68.00} & \cellcolor[HTML]{9DC8B6}{\color[HTML]{3B3B3B} 81.43} & \cellcolor[HTML]{A2CBBA}{\color[HTML]{3B3B3B} 77.14} & \cellcolor[HTML]{98C5B2}{\color[HTML]{3B3B3B} 85.71} & \cellcolor[HTML]{B7D6C9}{\color[HTML]{3B3B3B} 60.00} & \cellcolor[HTML]{E9F2EE}{\color[HTML]{3B3B3B} 18.18} & \cellcolor[HTML]{DBEBE4}{\color[HTML]{3B3B3B} 29.41} & \cellcolor[HTML]{A9CFBF}{\color[HTML]{3B3B3B} 71.43} & \cellcolor[HTML]{ADD1C2}{\color[HTML]{3B3B3B} 68.00} & \cellcolor[HTML]{9EC9B7}{\color[HTML]{3B3B3B} 80.49} & \cellcolor[HTML]{A2CBBA}{\color[HTML]{3B3B3B} 76.78} & \cellcolor[HTML]{9CC8B5}{\color[HTML]{3B3B3B} 82.04} & \cellcolor[HTML]{B7D6C9}{\color[HTML]{3B3B3B} 60.00} & \cellcolor[HTML]{F4F9F7}{\color[HTML]{3B3B3B} 8.74} & \cellcolor[HTML]{DBEBE4}{\color[HTML]{3B3B3B} 29.41} & \cellcolor[HTML]{D5E7E0}{\color[HTML]{3B3B3B} 34.34} \\
Selected & Phind-CodeLlama-34B-v2 & \cellcolor[HTML]{8BBEA9}{\color[HTML]{3B3B3B} 96.00} & \cellcolor[HTML]{94C3B0}{\color[HTML]{3B3B3B} 88.57} & \cellcolor[HTML]{91C1AD}{\color[HTML]{3B3B3B} 91.43} & \cellcolor[HTML]{9FC9B7}{\color[HTML]{3B3B3B} 80.00} & \cellcolor[HTML]{97C5B2}{\color[HTML]{3B3B3B} 86.00} & \cellcolor[HTML]{ACD1C1}{\color[HTML]{3B3B3B} 68.57} & \cellcolor[HTML]{9FC9B7}{\color[HTML]{3B3B3B} 80.00} & \cellcolor[HTML]{BAD8CC}{\color[HTML]{3B3B3B} 57.14} & \cellcolor[HTML]{9FC9B7}{\color[HTML]{3B3B3B} 80.00} & \cellcolor[HTML]{FFFFFF}{\color[HTML]{3B3B3B} 0.00} & \cellcolor[HTML]{B4D5C7}{\color[HTML]{3B3B3B} 62.35} & \cellcolor[HTML]{E7F1ED}{\color[HTML]{3B3B3B} 20.00} & \cellcolor[HTML]{98C5B2}{\color[HTML]{3B3B3B} 85.81} & \cellcolor[HTML]{ADD1C2}{\color[HTML]{3B3B3B} 67.54} & \cellcolor[HTML]{9FC9B8}{\color[HTML]{3B3B3B} 79.66} & \cellcolor[HTML]{BDDACE}{\color[HTML]{3B3B3B} 54.58} & \cellcolor[HTML]{B9D8CB}{\color[HTML]{3B3B3B} 58.06} & \cellcolor[HTML]{FFFFFF}{\color[HTML]{3B3B3B} 0.00} & \cellcolor[HTML]{B4D5C7}{\color[HTML]{3B3B3B} 62.32} & \cellcolor[HTML]{F4F9F7}{\color[HTML]{3B3B3B} 8.42} \\
Selected & gpt-3.5-turbo-1106 & \cellcolor[HTML]{93C2AE}{\color[HTML]{3B3B3B} 90.00} & \cellcolor[HTML]{99C6B3}{\color[HTML]{3B3B3B} 84.29} & \cellcolor[HTML]{87BCA6}{\color[HTML]{3B3B3B} 100.00} & \cellcolor[HTML]{8DBFAB}{\color[HTML]{3B3B3B} 94.29} & \cellcolor[HTML]{97C5B2}{\color[HTML]{3B3B3B} 86.00} & \cellcolor[HTML]{A2CBBA}{\color[HTML]{3B3B3B} 77.14} & \cellcolor[HTML]{99C6B3}{\color[HTML]{3B3B3B} 84.29} & \cellcolor[HTML]{A5CDBC}{\color[HTML]{3B3B3B} 74.29} & \cellcolor[HTML]{A5CCBC}{\color[HTML]{3B3B3B} 75.00} & \cellcolor[HTML]{FFFFFF}{\color[HTML]{3B3B3B} 0.00} & \cellcolor[HTML]{AACFC0}{\color[HTML]{3B3B3B} 70.59} & \cellcolor[HTML]{EAF3EF}{\color[HTML]{3B3B3B} 17.14} & \cellcolor[HTML]{97C5B2}{\color[HTML]{3B3B3B} 85.87} & \cellcolor[HTML]{A3CCBB}{\color[HTML]{3B3B3B} 75.98} & \cellcolor[HTML]{9AC6B4}{\color[HTML]{3B3B3B} 83.90} & \cellcolor[HTML]{A8CEBE}{\color[HTML]{3B3B3B} 72.24} & \cellcolor[HTML]{B4D5C7}{\color[HTML]{3B3B3B} 62.29} & \cellcolor[HTML]{FFFFFF}{\color[HTML]{3B3B3B} 0.00} & \cellcolor[HTML]{AACFC0}{\color[HTML]{3B3B3B} 70.59} & \cellcolor[HTML]{F8FBFA}{\color[HTML]{3B3B3B} 5.40} \\

\hline
\multicolumn{2}{r|}{\cellcolor[HTML]{FFFFFF}\textbf{Average (\textit{Selected})}} & \multicolumn{4}{c||}{{\color[HTML]{3B3B3B} \textbf{91.73}}} & \multicolumn{4}{c||}{{\color[HTML]{3B3B3B} \textbf{72.33}}} & \multicolumn{4}{c||}{{\color[HTML]{3B3B3B} \textbf{34.95}}} & \multicolumn{4}{c||}{{\color[HTML]{3B3B3B} \textbf{70.92}}} & \multicolumn{4}{c}{\cellcolor[HTML]{FFFFFF}{\color[HTML]{3B3B3B} \textbf{27.40}}} \\
\hline

Maximum & WizardCoder-15B-V1.0 & \cellcolor[HTML]{89BDA7}{\color[HTML]{3B3B3B} 98.00} & \cellcolor[HTML]{ABD0C0}{\color[HTML]{3B3B3B} 70.00} & \cellcolor[HTML]{94C3B0}{\color[HTML]{3B3B3B} 88.57} & \cellcolor[HTML]{BDDACE}{\color[HTML]{3B3B3B} 54.29} & \cellcolor[HTML]{B2D4C6}{\color[HTML]{3B3B3B} 64.00} & \cellcolor[HTML]{B8D7CA}{\color[HTML]{3B3B3B} 58.57} & \cellcolor[HTML]{B8D7CA}{\color[HTML]{3B3B3B} 58.57} & \cellcolor[HTML]{EDF5F2}{\color[HTML]{3B3B3B} 14.29} & \cellcolor[HTML]{D2E5DD}{\color[HTML]{3B3B3B} 37.50} & \cellcolor[HTML]{FFFFFF}{\color[HTML]{3B3B3B} 0.00} & \cellcolor[HTML]{E5F0EC}{\color[HTML]{3B3B3B} 21.18} & \cellcolor[HTML]{FFFFFF}{\color[HTML]{3B3B3B} 0.00} & \cellcolor[HTML]{B2D4C6}{\color[HTML]{3B3B3B} 63.97} & \cellcolor[HTML]{B9D8CB}{\color[HTML]{3B3B3B} 57.66} & \cellcolor[HTML]{B9D8CB}{\color[HTML]{3B3B3B} 57.85} & \cellcolor[HTML]{EFF6F3}{\color[HTML]{3B3B3B} 13.29} & \cellcolor[HTML]{D6E8E0}{\color[HTML]{3B3B3B} 34.12} & \cellcolor[HTML]{FFFFFF}{\color[HTML]{3B3B3B} 0.00} & \cellcolor[HTML]{E5F0EC}{\color[HTML]{3B3B3B} 21.18} & \cellcolor[HTML]{FFFFFF}{\color[HTML]{3B3B3B} 0.00} \\
Maximum & deepseek-coder-6.7b-instruct & \cellcolor[HTML]{8EC0AB}{\color[HTML]{3B3B3B} 94.00} & \cellcolor[HTML]{91C1AD}{\color[HTML]{3B3B3B} 91.43} & \cellcolor[HTML]{87BCA6}{\color[HTML]{3B3B3B} 100.00} & \cellcolor[HTML]{94C3B0}{\color[HTML]{3B3B3B} 88.57} & \cellcolor[HTML]{9FC9B7}{\color[HTML]{3B3B3B} 80.00} & \cellcolor[HTML]{ABD0C0}{\color[HTML]{3B3B3B} 70.00} & \cellcolor[HTML]{A9CFBF}{\color[HTML]{3B3B3B} 71.43} & \cellcolor[HTML]{9FC9B7}{\color[HTML]{3B3B3B} 80.00} & \cellcolor[HTML]{A5CCBC}{\color[HTML]{3B3B3B} 75.00} & \cellcolor[HTML]{FFFFFF}{\color[HTML]{3B3B3B} 0.00} & \cellcolor[HTML]{DBEBE4}{\color[HTML]{3B3B3B} 29.41} & \cellcolor[HTML]{ACD1C1}{\color[HTML]{3B3B3B} 68.57} & \cellcolor[HTML]{9FC9B7}{\color[HTML]{3B3B3B} 80.00} & \cellcolor[HTML]{ACD1C2}{\color[HTML]{3B3B3B} 68.38} & \cellcolor[HTML]{A9CFBF}{\color[HTML]{3B3B3B} 71.24} & \cellcolor[HTML]{ADD1C2}{\color[HTML]{3B3B3B} 68.06} & \cellcolor[HTML]{ACD0C1}{\color[HTML]{3B3B3B} 69.12} & \cellcolor[HTML]{FFFFFF}{\color[HTML]{3B3B3B} 0.00} & \cellcolor[HTML]{DBEBE4}{\color[HTML]{3B3B3B} 29.41} & \cellcolor[HTML]{EEF6F3}{\color[HTML]{3B3B3B} 13.39} \\
Maximum & deepseek-coder-33b-instruct & \cellcolor[HTML]{87BCA6}{\color[HTML]{3B3B3B} 100.00} & \cellcolor[HTML]{8CBEA9}{\color[HTML]{3B3B3B} 95.71} & \cellcolor[HTML]{87BCA6}{\color[HTML]{3B3B3B} 100.00} & \cellcolor[HTML]{87BCA6}{\color[HTML]{3B3B3B} 100.00} & \cellcolor[HTML]{A6CDBD}{\color[HTML]{3B3B3B} 74.00} & \cellcolor[HTML]{9FC9B7}{\color[HTML]{3B3B3B} 80.00} & \cellcolor[HTML]{A4CCBB}{\color[HTML]{3B3B3B} 75.71} & \cellcolor[HTML]{ACD1C1}{\color[HTML]{3B3B3B} 68.57} & \cellcolor[HTML]{AED1C2}{\color[HTML]{3B3B3B} 67.50} & \cellcolor[HTML]{E9F2EE}{\color[HTML]{3B3B3B} 18.18} & \cellcolor[HTML]{C7E0D6}{\color[HTML]{3B3B3B} 45.88} & \cellcolor[HTML]{E0EDE8}{\color[HTML]{3B3B3B} 25.71} & \cellcolor[HTML]{A6CDBD}{\color[HTML]{3B3B3B} 74.00} & \cellcolor[HTML]{9FC9B8}{\color[HTML]{3B3B3B} 79.36} & \cellcolor[HTML]{A4CCBB}{\color[HTML]{3B3B3B} 75.63} & \cellcolor[HTML]{B4D5C7}{\color[HTML]{3B3B3B} 62.23} & \cellcolor[HTML]{AED1C2}{\color[HTML]{3B3B3B} 67.50} & \cellcolor[HTML]{EEF5F2}{\color[HTML]{3B3B3B} 13.64} & \cellcolor[HTML]{C8E0D6}{\color[HTML]{3B3B3B} 45.67} & \cellcolor[HTML]{F2F7F5}{\color[HTML]{3B3B3B} 10.72} \\
Maximum & Phind-CodeLlama-34B-v2 & \cellcolor[HTML]{87BCA6}{\color[HTML]{3B3B3B} 100.00} & \cellcolor[HTML]{8FC0AC}{\color[HTML]{3B3B3B} 92.86} & \cellcolor[HTML]{8ABDA8}{\color[HTML]{3B3B3B} 97.14} & \cellcolor[HTML]{94C3B0}{\color[HTML]{3B3B3B} 88.57} & \cellcolor[HTML]{A3CCBB}{\color[HTML]{3B3B3B} 76.00} & \cellcolor[HTML]{B3D4C7}{\color[HTML]{3B3B3B} 62.86} & \cellcolor[HTML]{A5CDBC}{\color[HTML]{3B3B3B} 74.29} & \cellcolor[HTML]{BDDACE}{\color[HTML]{3B3B3B} 54.29} & \cellcolor[HTML]{ABD0C0}{\color[HTML]{3B3B3B} 70.00} & \cellcolor[HTML]{FFFFFF}{\color[HTML]{3B3B3B} 0.00} & \cellcolor[HTML]{B8D7CA}{\color[HTML]{3B3B3B} 58.82} & \cellcolor[HTML]{E7F1ED}{\color[HTML]{3B3B3B} 20.00} & \cellcolor[HTML]{A3CCBB}{\color[HTML]{3B3B3B} 76.00} & \cellcolor[HTML]{B5D5C8}{\color[HTML]{3B3B3B} 61.62} & \cellcolor[HTML]{A6CDBD}{\color[HTML]{3B3B3B} 74.09} & \cellcolor[HTML]{C2DDD1}{\color[HTML]{3B3B3B} 50.62} & \cellcolor[HTML]{B2D4C6}{\color[HTML]{3B3B3B} 64.02} & \cellcolor[HTML]{FFFFFF}{\color[HTML]{3B3B3B} 0.00} & \cellcolor[HTML]{B8D7CA}{\color[HTML]{3B3B3B} 58.82} & \cellcolor[HTML]{F5F9F8}{\color[HTML]{3B3B3B} 7.69} \\
Maximum & gpt-3.5-turbo-1106 & \cellcolor[HTML]{8EC0AB}{\color[HTML]{3B3B3B} 94.00} & \cellcolor[HTML]{98C5B2}{\color[HTML]{3B3B3B} 85.71} & \cellcolor[HTML]{87BCA6}{\color[HTML]{3B3B3B} 100.00} & \cellcolor[HTML]{8DBFAB}{\color[HTML]{3B3B3B} 94.29} & \cellcolor[HTML]{ADD1C2}{\color[HTML]{3B3B3B} 68.00} & \cellcolor[HTML]{ACD1C1}{\color[HTML]{3B3B3B} 68.57} & \cellcolor[HTML]{B1D3C5}{\color[HTML]{3B3B3B} 64.29} & \cellcolor[HTML]{B3D4C7}{\color[HTML]{3B3B3B} 62.86} & \cellcolor[HTML]{BDDACE}{\color[HTML]{3B3B3B} 55.00} & \cellcolor[HTML]{FFFFFF}{\color[HTML]{3B3B3B} 0.00} & \cellcolor[HTML]{D3E6DE}{\color[HTML]{3B3B3B} 36.47} & \cellcolor[HTML]{E3EFEA}{\color[HTML]{3B3B3B} 22.86} & \cellcolor[HTML]{ADD1C2}{\color[HTML]{3B3B3B} 67.56} & \cellcolor[HTML]{AFD2C4}{\color[HTML]{3B3B3B} 66.07} & \cellcolor[HTML]{B2D4C6}{\color[HTML]{3B3B3B} 63.88} & \cellcolor[HTML]{B7D6C9}{\color[HTML]{3B3B3B} 59.77} & \cellcolor[HTML]{C1DCD1}{\color[HTML]{3B3B3B} 51.47} & \cellcolor[HTML]{FFFFFF}{\color[HTML]{3B3B3B} 0.00} & \cellcolor[HTML]{D3E6DE}{\color[HTML]{3B3B3B} 36.47} & \cellcolor[HTML]{F5F9F8}{\color[HTML]{3B3B3B} 7.78} \\
\hline
\multicolumn{2}{r|}{\cellcolor[HTML]{FFFFFF}\textbf{Average (\textit{Maximum})}} & \multicolumn{4}{c||}{{\color[HTML]{3B3B3B} 91.66}} & \multicolumn{4}{c||}{{\color[HTML]{3B3B3B} 66.31}} & \multicolumn{4}{c||}{{\color[HTML]{3B3B3B} 32.60}} & \multicolumn{4}{c||}{{\color[HTML]{3B3B3B} 64.56}} & \multicolumn{4}{c}{{\color[HTML]{3B3B3B} 26.55}} \\
\hline
Minimum & WizardCoder-15B-V1.0 & \cellcolor[HTML]{87BCA6}{\color[HTML]{3B3B3B} 100.00} & \cellcolor[HTML]{A0CAB9}{\color[HTML]{3B3B3B} 78.57} & \cellcolor[HTML]{93C2AE}{\color[HTML]{3B3B3B} 90.00} & \cellcolor[HTML]{A2CBBA}{\color[HTML]{3B3B3B} 77.14} & \cellcolor[HTML]{E9F2EE}{\color[HTML]{3B3B3B} 18.00} & \cellcolor[HTML]{DBEAE4}{\color[HTML]{3B3B3B} 30.00} & \cellcolor[HTML]{C4DED3}{\color[HTML]{3B3B3B} 48.57} & \cellcolor[HTML]{FFFFFF}{\color[HTML]{3B3B3B} 0.00} & \cellcolor[HTML]{FFFFFF}{\color[HTML]{3B3B3B} 0.00} & \cellcolor[HTML]{FFFFFF}{\color[HTML]{3B3B3B} 0.00} & \cellcolor[HTML]{F7FBF9}{\color[HTML]{3B3B3B} 5.88} & \cellcolor[HTML]{FFFFFF}{\color[HTML]{3B3B3B} 0.00} & \cellcolor[HTML]{E9F2EE}{\color[HTML]{3B3B3B} 18.00} & \cellcolor[HTML]{DBEBE4}{\color[HTML]{3B3B3B} 29.22} & \cellcolor[HTML]{C4DED3}{\color[HTML]{3B3B3B} 48.52} & \cellcolor[HTML]{FFFFFF}{\color[HTML]{3B3B3B} 0.00} & \cellcolor[HTML]{FFFFFF}{\color[HTML]{3B3B3B} 0.00} & \cellcolor[HTML]{FFFFFF}{\color[HTML]{3B3B3B} 0.00} & \cellcolor[HTML]{F7FBF9}{\color[HTML]{3B3B3B} 5.88} & \cellcolor[HTML]{FFFFFF}{\color[HTML]{3B3B3B} 0.00} \\
Minimum & deepseek-coder-6.7b-instruct & \cellcolor[HTML]{9FC9B7}{\color[HTML]{3B3B3B} 80.00} & \cellcolor[HTML]{8FC0AC}{\color[HTML]{3B3B3B} 92.86} & \cellcolor[HTML]{87BCA6}{\color[HTML]{3B3B3B} 100.00} & \cellcolor[HTML]{98C5B2}{\color[HTML]{3B3B3B} 85.71} & \cellcolor[HTML]{BBD9CD}{\color[HTML]{3B3B3B} 56.00} & \cellcolor[HTML]{ECF4F1}{\color[HTML]{3B3B3B} 15.71} & \cellcolor[HTML]{BAD8CC}{\color[HTML]{3B3B3B} 57.14} & \cellcolor[HTML]{E3EFEA}{\color[HTML]{3B3B3B} 22.86} & \cellcolor[HTML]{FFFFFF}{\color[HTML]{3B3B3B} 0.00} & \cellcolor[HTML]{FFFFFF}{\color[HTML]{3B3B3B} 0.00} & \cellcolor[HTML]{FFFFFF}{\color[HTML]{3B3B3B} 0.00} & \cellcolor[HTML]{FFFFFF}{\color[HTML]{3B3B3B} 0.00} & \cellcolor[HTML]{BBD9CD}{\color[HTML]{3B3B3B} 55.89} & \cellcolor[HTML]{ECF4F1}{\color[HTML]{3B3B3B} 15.69} & \cellcolor[HTML]{BAD8CC}{\color[HTML]{3B3B3B} 57.14} & \cellcolor[HTML]{E8F2ED}{\color[HTML]{3B3B3B} 19.14} & \cellcolor[HTML]{FFFFFF}{\color[HTML]{3B3B3B} 0.00} & \cellcolor[HTML]{FFFFFF}{\color[HTML]{3B3B3B} 0.00} & \cellcolor[HTML]{FFFFFF}{\color[HTML]{3B3B3B} 0.00} & \cellcolor[HTML]{FFFFFF}{\color[HTML]{3B3B3B} 0.00} \\
Minimum & deepseek-coder-33b-instruct & \cellcolor[HTML]{87BCA6}{\color[HTML]{3B3B3B} 100.00} & \cellcolor[HTML]{91C1AD}{\color[HTML]{3B3B3B} 91.43} & \cellcolor[HTML]{88BCA7}{\color[HTML]{3B3B3B} 98.57} & \cellcolor[HTML]{9BC7B5}{\color[HTML]{3B3B3B} 82.86} & \cellcolor[HTML]{C5DED4}{\color[HTML]{3B3B3B} 48.00} & \cellcolor[HTML]{E3EFEA}{\color[HTML]{3B3B3B} 22.86} & \cellcolor[HTML]{B0D2C4}{\color[HTML]{3B3B3B} 65.71} & \cellcolor[HTML]{F1F7F4}{\color[HTML]{3B3B3B} 11.43} & \cellcolor[HTML]{F0F6F3}{\color[HTML]{3B3B3B} 12.50} & \cellcolor[HTML]{FFFFFF}{\color[HTML]{3B3B3B} 0.00} & \cellcolor[HTML]{D1E5DD}{\color[HTML]{3B3B3B} 37.65} & \cellcolor[HTML]{FFFFFF}{\color[HTML]{3B3B3B} 0.00} & \cellcolor[HTML]{C5DED4}{\color[HTML]{3B3B3B} 48.00} & \cellcolor[HTML]{E3EFEA}{\color[HTML]{3B3B3B} 22.58} & \cellcolor[HTML]{B1D3C5}{\color[HTML]{3B3B3B} 64.98} & \cellcolor[HTML]{F8FBFA}{\color[HTML]{3B3B3B} 5.21} & \cellcolor[HTML]{F0F6F3}{\color[HTML]{3B3B3B} 12.50} & \cellcolor[HTML]{FFFFFF}{\color[HTML]{3B3B3B} 0.00} & \cellcolor[HTML]{D2E5DD}{\color[HTML]{3B3B3B} 37.43} & \cellcolor[HTML]{FFFFFF}{\color[HTML]{3B3B3B} 0.00} \\
Minimum & Phind-CodeLlama-34B-v2 & \cellcolor[HTML]{8BBEA9}{\color[HTML]{3B3B3B} 96.00} & \cellcolor[HTML]{93C2AE}{\color[HTML]{3B3B3B} 90.00} & \cellcolor[HTML]{8ABDA8}{\color[HTML]{3B3B3B} 97.14} & \cellcolor[HTML]{94C3B0}{\color[HTML]{3B3B3B} 88.57} & \cellcolor[HTML]{AFD2C4}{\color[HTML]{3B3B3B} 66.00} & \cellcolor[HTML]{C6DFD5}{\color[HTML]{3B3B3B} 47.14} & \cellcolor[HTML]{BAD8CC}{\color[HTML]{3B3B3B} 57.14} & \cellcolor[HTML]{F4F9F7}{\color[HTML]{3B3B3B} 8.57} & \cellcolor[HTML]{F6F9F8}{\color[HTML]{3B3B3B} 7.50} & \cellcolor[HTML]{FFFFFF}{\color[HTML]{3B3B3B} 0.00} & \cellcolor[HTML]{FFFFFF}{\color[HTML]{3B3B3B} 0.00} & \cellcolor[HTML]{FFFFFF}{\color[HTML]{3B3B3B} 0.00} & \cellcolor[HTML]{B0D2C4}{\color[HTML]{3B3B3B} 65.69} & \cellcolor[HTML]{C7DFD5}{\color[HTML]{3B3B3B} 46.33} & \cellcolor[HTML]{BAD8CC}{\color[HTML]{3B3B3B} 57.14} & \cellcolor[HTML]{F5F9F7}{\color[HTML]{3B3B3B} 8.02} & \cellcolor[HTML]{F6F9F8}{\color[HTML]{3B3B3B} 7.50} & \cellcolor[HTML]{FFFFFF}{\color[HTML]{3B3B3B} 0.00} & \cellcolor[HTML]{FFFFFF}{\color[HTML]{3B3B3B} 0.00} & \cellcolor[HTML]{FFFFFF}{\color[HTML]{3B3B3B} 0.00} \\

Minimum & gpt-3.5-turbo-1106 & \cellcolor[HTML]{89BDA7}{\color[HTML]{3B3B3B} 98.00} & \cellcolor[HTML]{98C5B2}{\color[HTML]{3B3B3B} 85.71} & \cellcolor[HTML]{8CBEA9}{\color[HTML]{3B3B3B} 95.71} & \cellcolor[HTML]{91C1AD}{\color[HTML]{3B3B3B} 91.43} & \cellcolor[HTML]{AFD2C4}{\color[HTML]{3B3B3B} 66.00} & \cellcolor[HTML]{CFE4DB}{\color[HTML]{3B3B3B} 40.00} & \cellcolor[HTML]{B3D4C7}{\color[HTML]{3B3B3B} 62.86} & \cellcolor[HTML]{E7F1ED}{\color[HTML]{3B3B3B} 20.00} & \cellcolor[HTML]{F0F6F3}{\color[HTML]{3B3B3B} 12.50} & \cellcolor[HTML]{FFFFFF}{\color[HTML]{3B3B3B} 0.00} & \cellcolor[HTML]{FFFFFF}{\color[HTML]{3B3B3B} 0.00} & \cellcolor[HTML]{FFFFFF}{\color[HTML]{3B3B3B} 0.00} & \cellcolor[HTML]{AFD2C4}{\color[HTML]{3B3B3B} 65.94} & \cellcolor[HTML]{CFE4DB}{\color[HTML]{3B3B3B} 39.85} & \cellcolor[HTML]{B3D4C7}{\color[HTML]{3B3B3B} 62.86} & \cellcolor[HTML]{E7F2ED}{\color[HTML]{3B3B3B} 19.27} & \cellcolor[HTML]{F0F6F3}{\color[HTML]{3B3B3B} 12.50} & \cellcolor[HTML]{FFFFFF}{\color[HTML]{3B3B3B} 0.00} & \cellcolor[HTML]{FFFFFF}{\color[HTML]{3B3B3B} 0.00} & \cellcolor[HTML]{FFFFFF}{\color[HTML]{3B3B3B} 0.00} \\
\hline
\multicolumn{2}{r|}{\cellcolor[HTML]{FFFFFF}\textbf{Average (\textit{Minimum})}} & \multicolumn{4}{c||}{{\color[HTML]{3B3B3B} 90.99}} & \multicolumn{4}{c||}{{\color[HTML]{3B3B3B} 38.20}} & \multicolumn{4}{c||}{{\color[HTML]{3B3B3B} 3.80}} & \multicolumn{4}{c||}{{\color[HTML]{3B3B3B} 37.47}} & \multicolumn{4}{c}{{\color[HTML]{3B3B3B} 3.79}} \\
\hline
\multicolumn{2}{r|}{\textbf{Overall Average}} & \multicolumn{4}{c}{\cellcolor[HTML]{FFFFFF}\textbf{91.46}} & \multicolumn{4}{c}{\cellcolor[HTML]{FFFFFF}\textbf{58.95}} & \multicolumn{4}{c}{\cellcolor[HTML]{FFFFFF}\textbf{23.78}} & \multicolumn{4}{c}{\cellcolor[HTML]{FFFFFF}\textbf{57.65}} & \multicolumn{4}{c}{\cellcolor[HTML]{FFFFFF}\textbf{19.25}} \\
\bottomrule

\end{tabular}
}
\end{table*}

\begin{figure}[th]
    \centering
    \includegraphics[width=1.0\linewidth]{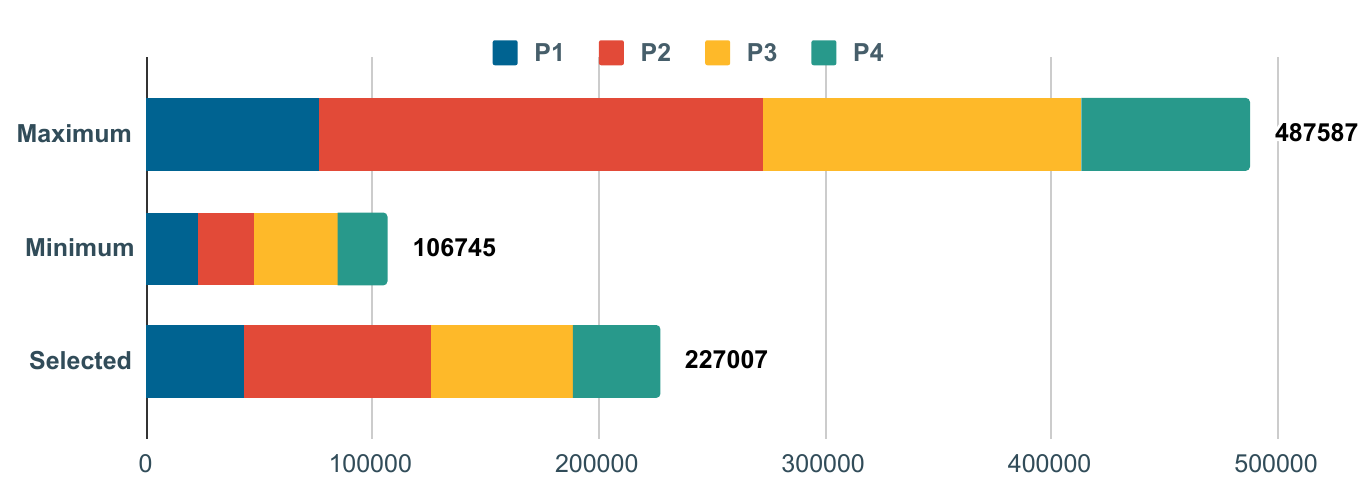}
    \setlength{\abovecaptionskip}{-4pt}
    \setlength{\belowcaptionskip}{-10pt}
    \caption{Number of Characters of Three Context Settings (\ie, Maximum/Minimum and Selected Context, Section~\ref{sec:context}). Each color represents each project in \name.}
    \label{fig:token}
\end{figure}

In RQ1, the context setting is fixed to the \textit{selected context}, where only the context that is related to the class/function to be generated is fed into LLMs. In RQ2, we consider the other two context settings (\ie, maximum and minimum context). We fix the synthesis strategy as holistic because it performs the best in RQ1. The experiment result of RQ2 is visualized in Table~\ref{tab:rq2}. 

From Table~\ref{tab:rq2}, it is clear that among three context settings, the \textbf{\textit{selected context}} yield the best overall results, with 70.92\% (class-wise) and 27.40\% (test-wise) Pass@1, which echos the results in Table~\ref{tab:rq1}. Though maximum and minimum context achieve similar Completion@1 (\ie, 91.66\% and 90.99\% compared with 91.73\%), 
the performance of using these two contexts on the subsequent metrics (\ie, Compilation@1 and Pass@1) is not as good as the selected context.

\begin{mdframed}[style=MyFrame]
\textbf{Finding 5: }The \textbf{\textit{selected context is the best setting}} on \name, resulting in \textbf{\textit{70.92\%}} (class-wise) Pass@1. 
\end{mdframed}

To better understand the size of the total context used by each setting, we visualize the \textbf{\textit{number of characters}} under each setting (\ie, \textit{maximum}, \textit{minimum} and \textit{selected} context) in Figure~\ref{fig:token}. 
Four bars in each group are P1-P4 in \name. Note that we calculate the number of characters instead of the tokens because LLMs utilize different tokenizers, which will affect the counts of tokens. 
For example, WizardCoder uses GPT2Tokenizer, Phind uses LlamaTokenizer, while DeepSeek uses LlamaTokenizerFast. From Figure~\ref{fig:token}, it is clear that the number of characters used in \textit{Maximum Context} is nearly five times that of \textit{Minimum Context} and more than twice that of \textit{Selected Context}. Additionally, we can observe that the number of characters for each project is relatively similar, with P2 and P3 having relatively more characters and P4 having fewer.

By combining Table~\ref{tab:rq2} and Figure~\ref{fig:token}, we can make two observations. First, \textbf{\textit{more context is not always a benefit.}}
For example, a dramatic 54.83\% (= 69.12\%-14.29\%) \textbf{\textit{increase}} in test-wise Pass@1 is achieved by DeepSeek-6.7b when switching the context from \textit{selected} to \textit{maximum} context in P1, while a 23.62\% (= 34.34\%-10.72\%) \textbf{\textit{drop}} of test-wise Pass@1 in P4 can be observed in DeepSeek-33b. 
Second, \textbf{\textit{only providing the class to be completed is insufficient}} without dependencies. We can see from Table~\ref{tab:rq2} that the Pass@1 in the test-wise granularity is \textbf{\textit{almost all zeros}} under the minimum context setting, meaning that it is almost impossible to generate project-level code that can pass test cases. In addition, the \textbf{\textit{selected context}} includes only the method signatures (as explained in Section~\ref{sec:context}), which turned out to be more effective than the maximum context setting. 

\begin{mdframed}[style=MyFrame]
\textbf{Finding 6: }Providing {{too much or too little context has a negative impact on project-level }}code generation. \textbf{\textit{Including the method signature only}} shows a promising performance.
\end{mdframed}

\subsection{RQ3: Impact of Incremental Synthesis}

Since in Table~\ref{tab:rq1}, we only adopt \textbf{\textit{sequential}} order to synthesize functions incrementally. We then further explore whether the order of synthesizing functions in the class matters. Due to the space limitation, we only show the impact of DeepSeek-Coder-6.7b on Completion/Compilation/Pass@1 and @5. Similar observations are made in other LLMs, found in our released artifact.

\begin{figure}[t]
    \centering
    \includegraphics[width=1.0\linewidth]{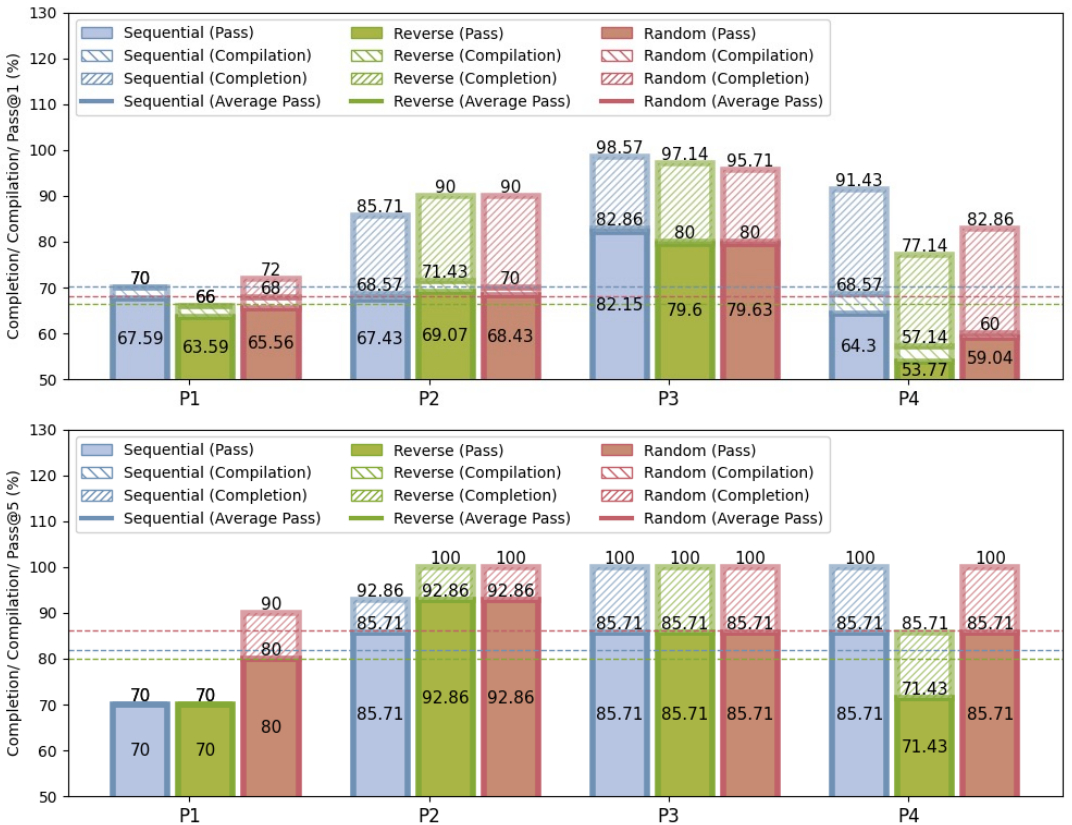}
    \setlength{\abovecaptionskip}{-5pt}
    \setlength{\belowcaptionskip}{-15pt}
    \caption{RQ3: Impact of Different Incremental Synthesis on DeepSeek-Coder-6.7b. Completion/Compilation/Pass@1 (Upper) and Completion/Compilation/Pass@5 (Lower). 
    }
    \label{fig:rq3}
\end{figure}

The result on four projects in \name is illustrated in Figure~\ref{fig:rq3}. 
The dashed horizontal lines in particular colors show the average across four projects.
Generally, the three colored bars in each project are similar, with slight variances among projects, meaning that the order of incremental synthesis can slightly affect the final results. Interestingly, with sampling size \textit{k} increases from 1 to 5, \textbf{the advantage of random order} is more significant. From the lower part of Figure~\ref{fig:rq3}, the red dashed line outperforms the other two lines, with an advantage of 6\% (86.07\% - 80\%) in the reversed order. On the other hand, the reversed order may not contribute to better results, according to the experiment. 

\begin{mdframed}[style=MyFrame]
\textbf{Finding 7: }Synthesizing programs incrementally in a \textbf{\textit{random}} and \textbf{\textit{sequential}} order could yield at most 6\% improvement than a reversed order on DeepSeek-Coder-6.7b. A similar conclusion was also observed in other studied LLMs.
\end{mdframed}

\subsection{RQ4: Bad Case Analysis}
This section discussed failures during completion/compilation/pass, showed the distribution of runtime error, and analyzed five bad cases that failed to compile and pass the test suites.

\subsubsection{\textbf{{Completion Errors}}} 
A completion error happens when LLMs ignore the code to be completed and leave the method body blank.
Similar observations (\ie, LLMs ignore the information in the middle of long contexts) were also made in other communities~\cite{liu2024lost}.
From Table~\ref{tab:rq1} and Table~\ref{tab:rq2} under `Completion@1', we can see that among the studied LLMs, completion errors were more commonly observed in WizardCoder than in others.

\subsubsection{\textbf{{Compilation Errors}}} 
Compilation errors indicate that LLMs have an insufficient understanding of the syntactic and semantic constraints provided by the context. Since Java's compilation errors are not explicitly categorized, we can only roughly determine the cause of each compilation error in the code through manual judgment. Below, we present three typical errors that occur frequently and are related to the object-oriented programming paradigm.

{\textbf{\textit{Compilation Error 1: \textbf{Finding 6: }Inheritance-related Error.}}} An abstract class is meant to be inherited by other classes and cannot be instantiated. 
However, in Listing~\ref{listing:ce1}, the \texttt{Move} class is defined as an abstract class in line 1, but it is instantiated in line 12, resulting in a compilation error.

\definecolor{stubbg}{HTML}{FCF3D5}

\begin{figure}[h!]
\begin{lstlisting}[
		language=java,
        % basicstyle=\linespread{0.8}\tiny,
        % baselinestretch=1.0,
		% morekeywords={var},
            aboveskip=-2pt,
            % belowskip=-15pt,
		caption={\textbf{Compilation Error 1. Inheritance-related Error}},
		label={listing:ce1},
		escapechar=|,
		linebackgroundcolor = {
                \ifnum \value{lstnumber} = 12 \color{removed} \fi 
                \ifnum \value{lstnumber} = 13 \color{added} \fi 
            },
		numbers=left
	]
public abstract class Move extends Action {}
public static final class Down extends Move {}

public class TerminalInputEngine implements InputEngine {
  @Override
  public Action fetchAction() {
    String actionName = ...;
    switch (actionName) {
      ...
      case "move":
        final Move.Direction direction = Move.Direction.valueOf(args.toUpperCase());
|\textbf{\textcolor{red}{-}}|       return new Move(initiator, direction);
|\textbf{\textcolor{teal}{+}}|       return new Move.Up(initiator);
    }
  }
}
\end{lstlisting}
\end{figure}

{\textbf{\textit{Compilation Error 2: Encapsulation-related Error}}}. Listing~\ref{listing:ce2} shows an error caused by improperly handled encapsulation.
The \texttt{player} field defined in line 3 is private in class \texttt{Piece} and cannot be accessed directly by its subclasses. LLMs ignore the principle of encapsulation and access the \texttt{player} field in line 13, causing this compilation error. 

\definecolor{stubbg}{HTML}{FCF3D5}

\begin{figure}[h!]
\begin{lstlisting}[
		language=java,
        % basicstyle=\linespread{0.8}\tiny,
        % baselinestretch=1.0,
		% morekeywords={var},
            aboveskip=-2pt,
            belowskip=-10pt,
		caption={\textbf{Compilation Error 2. Encapsulation-related Error}},
		label={listing:ce2},
		escapechar=|,
		linebackgroundcolor = {
                \ifnum \value{lstnumber} = 12 \color{removed} \fi
                \ifnum \value{lstnumber} = 13 \color{added} \fi
            },
		numbers=left
	]
public abstract class Piece {
  private final Player player;

  public final Player getPlayer() {	
    return this.player;	
  }	
}	

public class Knight extends Piece {	
  public Move[] getAvailableMoves(Game game, Place source) {	
    ...	
|\textbf{\textcolor{red}{-}}|   if (this.player.validateMove(game, move)) {
|\textbf{\textcolor{teal}{+}}|   if (this.getPlayer().validateMove(game, move)) {
      ...
    }	
  }	
}
\end{lstlisting}
\end{figure}

{\textbf{Compilation error 3: Illegal Inheritance}} Listing~\ref{listing:ce3} shows a case where LLMs fail to resolve the \textit{inheritance relationships}. The class \texttt{Player} defined in line 3 has no inheritance relationship with the \texttt{Cell} class defined in line 1, while LLMs mistakenly consider the variable \texttt{cell} could be an instance of \texttt{Player} using the \texttt{instanceof} keyword in line 10, causing the compilation fail.

\definecolor{stubbg}{HTML}{FCF3D5}

\begin{figure}[ht]
\begin{lstlisting}[
		language=java,
        % basicstyle=\linespread{0.8}\tiny,
        % baselinestretch=1.0,
		% morekeywords={var},
            aboveskip=-2pt,
            belowskip=-15pt,
		caption={\textbf{Compilation Error 3. Illegal Inheritance}},
		label={listing:ce3},
		escapechar=|,
		linebackgroundcolor = {
                \ifnum \value{lstnumber} = 10 \color{removed} \fi
                \ifnum \value{lstnumber} = 11 \color{added} \fi
                \ifnum \value{lstnumber} = 12 \color{added} \fi
                \ifnum \value{lstnumber} = 13 \color{added} \fi
            },
		numbers=left
	]
public abstract class Cell implements BoardElement {}
public abstract class Entity implements BoardElement {}
public final class Player extends Entity {}
public class EntityCell extends Cell {}

public class GameBoard {
  public Gameboard(...) {
    ...
    Cell cell = ...;
|\textbf{\textcolor{red}{-}}|   if (cell instanceof Player) {}
|\textbf{\textcolor{teal}{+}}|   if (cell instanceof EntityCell entityCell) {
|\textbf{\textcolor{teal}{+}}|     if (entityCell.getEntity() instanceof Player) {}
|\textbf{\textcolor{teal}{+}}|   }
    ...
  }
}
\end{lstlisting}
\end{figure}

\subsubsection{\textbf{{Test-Failing Errors}}}
\begin{figure}[t!]
    \centering
    \includegraphics[width=1.0\linewidth]{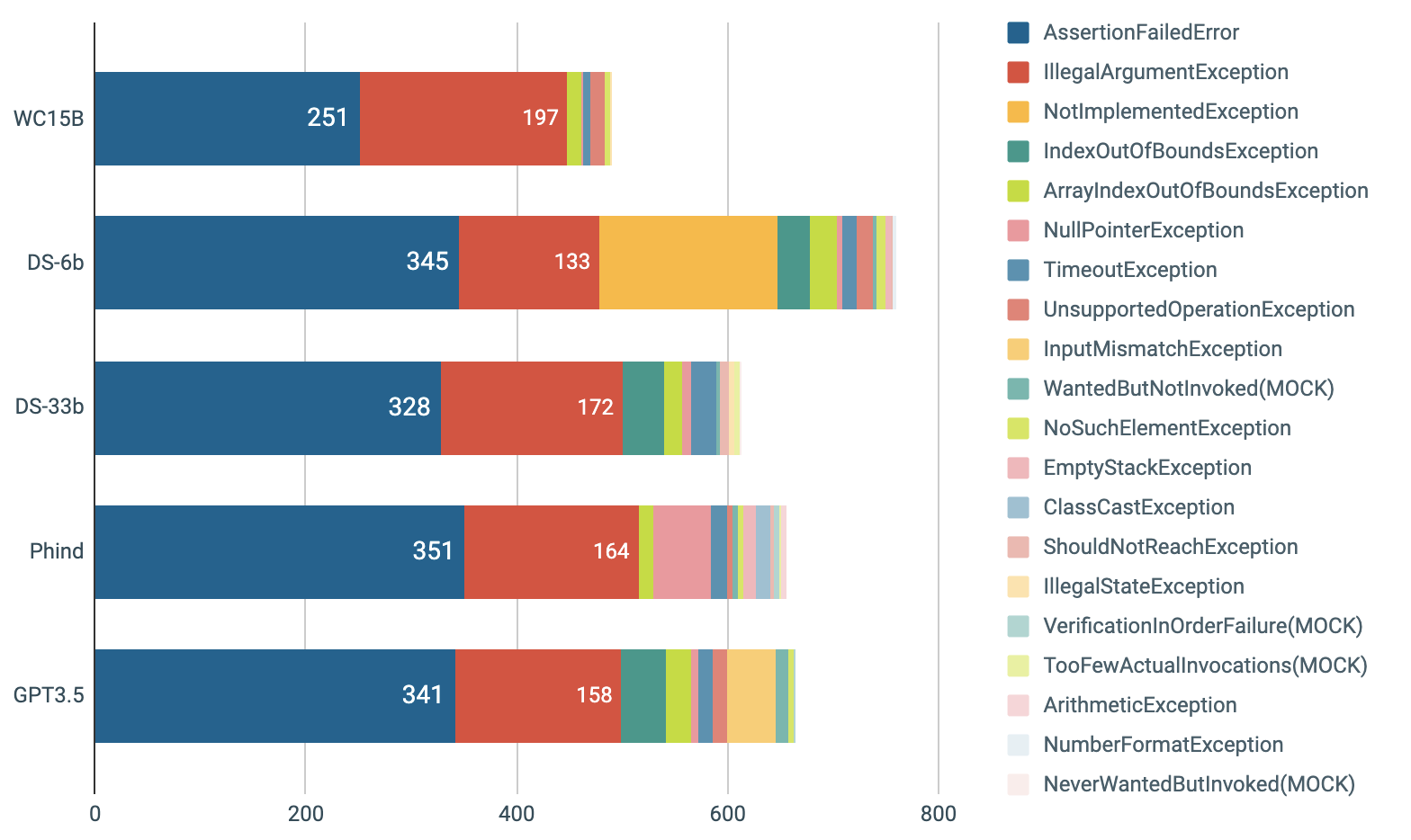}
    \setlength{\abovecaptionskip}{-2pt}
    \setlength{\belowcaptionskip}{-10pt}
    \caption{Exception Distribution in LLM-generated Code}
    \label{fig:error}
\end{figure}

The failure of test cases is often accompanied by exceptions thrown during execution.
We automatically parsed the error logs and presented the \textbf{\textit{exception distribution among LLMs}} in Figure~\ref{fig:error}.
Each stacked bar shows the exceptions thrown while running the projects synthesized by the corresponding LLM.
Different colors represent different types of exceptions.
In total, there are 20 types of test-failing errors. 
Among them, \texttt{AssertionFailedError} happens the most frequently (50.75\%), followed by \texttt{IllegatlAugumentException} (25.88\%). 

\begin{mdframed}[style=MyFrame]
\textbf{Finding 8: }\textbf{\texttt{AssertionFailedError}} and \textbf{\texttt{IllegalAugument-}}\\
\textbf{\texttt{Exception}} are \textbf{\textit{Top-2}} dominating contributors to test failures, accounting for 76.63\% test-failing errors.
\end{mdframed}

In particular, \texttt{AssertionFailedError} may indicate the LLMs did not fully understand the functionality written in the docstring so the assertions in the test case failed.
While \texttt{Illegal\-Argument\-Exception} mainly suggested a lack of understanding of the code constraints, leading to illegal arguments.
In the following, we analyzed two representative cases.

\textbf{\textit{Test-failing Error 1: Documentation Non-Following}.} 
In Listing~\ref{listing:te1}, the documentation of method \texttt{getUndoCount} in lines 12-14 clearly stated that \texttt{count} is the number of \texttt{pop()} calls. However, the method \texttt{push} generated by LLM also increases \texttt{count} in line 9, mistakenly understanding \texttt{count} as the size of \texttt{cellStack} and violating the documentation.

\definecolor{stubbg}{HTML}{FCF3D5}

\begin{figure}[hb]
\begin{lstlisting}[
		language=java,
        % basicstyle=\linespread{0.8}\tiny,
        % baselinestretch=1.0,
		% morekeywords={var},
            aboveskip=-10pt,
            belowskip=-8pt,
		caption={\textbf{Test Error 1. Documentation Non-Following}},
		label={listing:te1},
		escapechar=|,
  		linebackgroundcolor = {
                \ifnum \value{lstnumber} = 9 \color{removed} \fi
            },
		numbers=left
	]
public class CellStack {
  private final Stack<FillableCell> cellStack = new Stack<>();
  private int count = 0;
  
  void push(final FillableCell cell) {
    cellStack.push(cell);
    // count is the number of pop() invoked
    // rather than the size of cellStack.
|\textbf{\textcolor{red}{-}}|   count++;
  }
  
  /**
   * @return Number of undos (i.e. {@link CellStack#pop()}) invoked.
   */
  int getUndoCount() {
    return count;
  }
}
\end{lstlisting}
\end{figure}

\textbf{\textit{Test Error 2: Trivial Implementation.}}
In this case, the LLM produced a trivial implementation in line 5, constructing and returning a \texttt{Move} array of size zero.
This implementation is evidently a placeholder meant solely to pass compilation.
When the \texttt{Random::nextInt} method is called in line 12 and receives the parameter (i.e., the length of the Move array), its internal implementation first checks the parameter.
Upon finding it to be less than or equal to zero, it throws an \texttt{IllegalArgumentException}.

\definecolor{stubbg}{HTML}{FCF3D5}
\begin{figure}[th]
\begin{lstlisting}[
		language=java,
        % basicstyle=\linespread{0.8}\tiny,
        % baselinestretch=1.0,
		% morekeywords={var},
            aboveskip=-2pt,
            % belowskip=-20pt,
		caption={\textbf{Test Error 2. Trivial Implementation}},
		label={listing:te2},
		escapechar=|,
		% linebackgroundcolor = {\ifnum \value{lstnumber} > 2 \ifnum \value{lstnumber} < 10 \color{stubbg} \fi \fi},
		numbers=left
	]
public JesonMor extends Game {
  @Override
  public Move[] getAvailableMoves(Player player) {
    // This is a trivial implementation.
    return new Move[0];
  }
}

Move[] availableMoves = game.getAvailableMoves(player);
// Random::nextInt will throw IllegalArgumentException 
// if the input n <= 0.
int next = new Random().nextInt(availableMoves.length);
\end{lstlisting}
\end{figure}

\section{Threats to validity}

This paper has three main threats to validity.
First, \textit{\textbf{the threats in benchmark construction}}.
Also, the quality and detailed level of the natural language descriptions for the projects, classes, and methods could affect LLMs' code generation.
To alleviate this threat, we carefully checked the subjects in \name, scanned students' feedback, and adjusted the descriptions to mitigate confusion and ambiguity. 
Second, \textbf{\textit{the generalizability to other LLMs}}.
In this study, we only studied five LLMs due to time and hardware limits, so the conclusion may not be able to generalize to other LLMs.
Nonetheless, we selected the SOTA LLMs in different families as representatives (Section~\ref{subsec:llms}).
Third, \textbf{\textit{the performance variance brought by prompt engineering}}.
Since choosing one best-performing prompt is challenging~\cite{prompt-survey-2023}, and a well-designed prompt could yield better performance, so we follow the common practice of prompting LLMs~\cite{wizardcoder} to design the prompt template, trying to alleviate this threat.
Finally, \textbf{\textit{the possible data contamination}}~\cite{dataContamination2023} of \name. LLMs having seen the canonical code during training could lead to exaggerated scores, known as data contamination~\cite{dataContamination2023}.
However, the projects in \name were kept confidential (Section~\ref{sec:proj}), thus having minor concerns.

\section{Related Work}\label{sec:related}

\subsection{Benchmarks for Code Generation}

\subsubsection{Programming Language}
Most benchmarks target Python (Table~\ref{tab:benchmarks}), which is dynamic and rich in handy libraries, making it a good fit for building applications.
In contrast, Java is static, object-oriented, and with abundant design patterns, making it ideal for constructing large projects.
With such a different style from Python, Java, the most popular static language, receives far less attention in code generation benchmarks, especially at the project level.
\name is thus proposed to bridge the gap.

\subsubsection{Metric}
Similarity measurements such as exact match and BLEU~\cite{bleu} used to be mainstream metrics for code generation, as in other NLP tasks.
Recently, it was found that such similarity measurement has a weak correlation with semantic correctness~\cite{humaneval}.
More and more benchmarks~\cite{humaneval, du2023classeval, evalplus} adopt execution-based correctness such as Pass@$k$ as the major metric. 
Previous project-level benchmarks~\cite{repoeval, devEval} adopt a pass rate against test cases to assess LLM's capability to generate codes.
\name assesses LLMs with progressive metrics (completion to compilation to testing) and at finer granularities (class-wise and test-wise).

\subsubsection{Context}
The context of code generation benchmarks varies from line of code to project.
Function-level benchmarks such as HumanEval~\cite{humaneval} remain most commonly adopted because they are easy to evaluate and have certain discrimination.
As LLMs get more powerful, recent benchmarks target more complex contexts, such as class-level~\cite{du2023classeval} and project-level~\cite{repoeval, devEval, wang2024oop}.
Different from these benchmarks whose subjects were sourced from GitHub, \name is built from entry-level projects that are carefully designed to assess students' coding ability.
\name provides a straightforward comparison of LLMs' programming capability against humans.

\subsection{LLM-based Code Generation}

The code generation technique had a leap enabled by LLMs.
In particular, LLMs can handle a much more complex context (e.g., class or project) than symbolic-based approaches~\cite{flashfill11, feng17, kensen19} or small-sized language models~\cite{codebert, codet5}.

\subsubsection{Retrieval-Augmented Generation}
The code generation technique most relevant to our work is Retrieval-Augmented Generation (RAG).
A code project has a long context (documentation and codes), which usually exceeds LLMs' context limit or exposes performance degradation~\cite{liu2024lost}.
Therefore, a long context is usually decomposed into smaller chunks, and only the most related chunks are retrieved based on the problem and included in the prompt as context.
For example, RepoCoder~\cite{repoeval} uses a sliding window to decompose a project and does retrievals iteratively with LLM-generated codes.
Shrivastava et al.~\cite{Shrivastava23} has a finer project decomposition into different kinds (e.g., imported classes, child classes) and composes different information in the prompt.
\name adopts a simple but effective selected context setting that only includes the signatures of dependent types and can be adopted to complement RAG.

\section{Conclusion}\label{sec:conclusion}
In this paper, we introduce a project-level benchmark, \name, to fill the gap of the scarcity and need for high-quality Java benchmarks for LLM evaluation. 
Intensive experiments are conducted, covering the context settings, synthesis strategies, evaluation granularities, and evaluation metrics.

\section{Data Availability}\label{sec:data}
We released the implementation and all associated publicly available data at \url{https://github.com/java-bench/JavaBench}. 
We also release a {{leaderboard}} and invite model developers to participate and test their models against \name at \url{https://java-bench.github.io/leaderboard.html}.


\balance
\bibliographystyle{ACM-Reference-Format}
\bibliography{Tex/reference}



\end{document}